\documentclass[runningheads]{llncs}
\usepackage{graphicx}

\usepackage{tikz}
\usepackage{comment}
\usepackage{amsmath,amssymb} 
\usepackage{color}

\usepackage[accsupp]{axessibility}  
\usepackage{wrapfig}

\usepackage{graphicx}
\usepackage{subcaption}
\graphicspath{{./figures/}}
\usepackage{amsmath}
\usepackage{amssymb}
\usepackage{booktabs}
\newcommand{\norm}[1]{\left\lVert#1\right\rVert}
\usepackage{algorithm, setspace}
\usepackage{algpseudocode, array, multirow}
\usepackage{caption}
\usepackage{xcolor}
\usepackage{extarrows}
\usepackage{enumitem}
\usepackage{booktabs}
\usepackage{rotating}
\usepackage{fixltx2e}
\usepackage{textcomp}
\renewcommand{\textuparrow}{$\uparrow$}
\renewcommand{\textdownarrow}{$\downarrow$}
\definecolor{darkgreen}{rgb}{0.035, 0.412, 0.098}
\definecolor{mygreen}{rgb}{0.05, 0.71, 0.47}
\usepackage[breaklinks,colorlinks]{hyperref}
\hypersetup{colorlinks,breaklinks,urlcolor=[rgb]{0,0,0},linkcolor=[rgb]{0,0,0},citecolor=[rgb]{0,0,0}}
\usepackage[capitalize]{cleveref}
\crefname{section}{Sec.}{Secs.}
\crefname{table}{Tab.}{Tabs.}
\Crefname{equation}{Eq.}{Eqs.}
\Crefname{figure}{Fig.}{Figs.}
\Crefname{algorithm}{Alg.}{Algs.}

\newcommand{\etal}{\textit{et al.}}
\DeclareMathOperator*{\argmax}{arg\,max}
\DeclareMathOperator*{\argmin}{arg\,min}
\usepackage{orcidlink}

\begin{document}
\pagestyle{headings}
\mainmatter
\def\ECCVSubNumber{5478}  

\title{$\ell_\infty$-Robustness and Beyond: \\ Unleashing Efficient Adversarial Training} 

\titlerunning{$\ell_\infty$-Robustness and Beyond: Unleashing Efficient Adversarial Training}
%
\author{Hadi M.~Dolatabadi\textsuperscript{\orcidlink{0000-0001-9418-1487}} \and
Sarah Erfani\textsuperscript{\orcidlink{0000-0003-0885-0643}} \and
Christopher Leckie\textsuperscript{\orcidlink{0000-0002-4388-0517}}}
\authorrunning{H.~M.~Dolatabadi~\etal}
%
\institute{School of Computing and Information Systems\\The University of Melbourne\\Parkville, Victoria, Australia\\
\email{hadi.mohagheghdolatabadi@student.unimelb.edu.au}}
\maketitle

\begin{abstract}
 Neural networks are vulnerable to adversarial attacks: adding well-crafted, imperceptible perturbations to their input can modify their output.
 Adversarial training is one of the most effective approaches in training robust models against such attacks.
 However, it is much slower than vanilla training of neural networks since it needs to construct adversarial examples for the entire training data at every iteration, hampering its effectiveness.
 Recently, \textit{Fast Adversarial Training}~(FAT) was proposed that can obtain robust models efficiently.
 However, the reasons behind its success are not fully understood, and more importantly, it can only train robust models for $\ell_\infty$-bounded attacks as it uses FGSM during training.
 In this paper, by leveraging the theory of coreset selection, we show how selecting a small subset of training data provides a \textit{general}, more principled approach toward reducing the time complexity of robust training.
 Unlike existing methods, our approach can be adapted to a wide variety of training objectives, including TRADES, $\ell_p$-PGD, and Perceptual Adversarial Training~(PAT).
 Our experimental results indicate that our approach speeds up adversarial training by 2-3 times while experiencing a slight reduction in the clean and robust accuracy.
 
\end{abstract}

\section{Introduction}\label{sec:introduction}

Neural networks have achieved great success in the past decade.
Today, they are one of the primary candidates in solving a wide variety of machine learning tasks, from object detection and classification~\cite{he2016deep,wu2019detectron2} to photo-realistic image generation~\cite{karras2020stylegan2,vahdat2020nvae} and beyond.
Despite their impressive performance, neural networks are vulnerable to adversarial attacks~\cite{biggio2013evasion,szegedy2014intriguing}: adding well-crafted, imperceptible perturbations to their input can change their output.
This unexpected behavior of neural networks prevents their widespread deployment in safety-critical applications, including autonomous driving~\cite{eykholt2018robust} and medical diagnosis~\cite{ma2021understanding}.
As such, training robust neural networks against adversarial attacks is of paramount importance and has gained lots of attention.

\textit{Adversarial training} is one of the most successful approaches in defending neural networks against adversarial attacks.\footnote{Note that adversarial training in the literature generally refers to a particular approach proposed by Madry~\etal~\cite{madry2018towards}.
	For the purposes of this paper, we refer to any method that builds adversarial attacks around the training data and incorporates them into the training of the neural network as adversarial training. Using this taxonomy, methods such as TRADES~\cite{zhang2019trades}, $\ell_p$-PGD~\cite{madry2018towards} or Perceptual Adversarial Training~(PAT)~\cite{laidlaw2021pat} are all considered different versions of adversarial training.}
This approach first constructs a perturbed version of the training data.
Then, the neural network is optimized on these perturbed inputs instead of the clean samples.
This procedure must be done iteratively as the perturbations depend on the neural network weights.
Since the weights are optimized during training, the perturbations must also be adjusted for each data sample in every iteration.

Various adversarial training methods primarily differ in how they define and find the perturbed version of the input~\cite{madry2018towards,zhang2019trades,laidlaw2021pat}.
However, they all require repetitive construction of these perturbations during training which is often cast as another non-linear optimization problem.
As such, the time and computational complexity of adversarial training is massively higher than vanilla training.
In practice, neural networks require massive amounts of training data~\cite{adadi2021data} and need to be trained multiple times with various hyper-parameters to get their best performance~\cite{killamsetty2021gradmatch}.
Thus, reducing the time/computational complexity of adversarial training is critical in enabling the environmentally efficient application of robust neural networks in real-world scenarios~\cite{schwartz2020greenai,strubell2019energy}.

\textit{Fast Adversarial Training}~(FAT)~\cite{wong2020fast} is a successful approach proposed for efficient training of robust neural networks.
Contrary to the common belief that building the perturbed versions of the inputs using \textit{Fast Gradient Sign Method}~(FGSM)~\cite{goodfellow2014explaining} does not help in training arbitrary robust models~\cite{tramer2018ensemble,madry2018towards}, Wong~\etal~\cite{wong2020fast} show that by carefully applying uniformly random initialization before the FGSM step one can make this training approach work.
Using FGSM to generate the perturbed input in a single step combined with implementation tricks such as mixed precision and cyclic learning rate, FAT can significantly reduce the training time of robust neural networks.

Despite its success, FAT may exhibit unexpected behavior in different settings.
For instance, it was shown that FAT suffers from \textit{catastrophic overfitting} where the robust accuracy during training suddenly drops to 0\%~\cite{wong2020fast,andriushchenko2020understanding}.
A more fundamental issue with FAT and its variations such as \texttt{GradAlign}~\cite{andriushchenko2020understanding} is that they are specifically designed and implemented for $\ell_\infty$ adversarial training.
This is because FGSM, particularly an $\ell_\infty$ perturbation generator, is at the heart of these methods.
As a result, the quest for a unified, systematic approach that can reduce the time complexity of all types of adversarial training is not over.

Motivated by the limited scope of FAT, in this paper we take an important step towards finding a general yet principled approach for reducing the time complexity of adversarial training.
We notice that repetitive construction of adversarial examples for each data point is the main bottleneck of robust training.
While this process needs to be done iteratively, we speculate that perhaps we can find a subset of the training data that is more important to robust network optimization than the rest.
Specifically, we ask the following research question: \textit{Can we train an adversarially robust neural network using a subset of the entire training data without sacrificing clean or robust accuracy?}

\begin{figure*}[t!]
	\centering
	\begin{subfigure}{\textwidth}
		\centering
		\tikzstyle{box} = [rectangle, draw, fill=blue!20, text width=1.75cm, text badly centered, node distance=1cm, minimum height=.85cm, inner sep=0pt, rotate=-90]
		\tikzstyle{start} = [circle, draw, fill=green!20, text width=0.8cm, text centered, rounded corners, minimum height=0.6cm]
		\begin{tikzpicture}[node distance = 1cm, thick, scale=0.9, every node/.style={scale=0.75}]
		\node [start]              at (-0.5, 0.)  (start)  {Start};
		\node [box]                at (3., 0.)  (core1)  {Coreset\\Selection};
		\node [box]                at (6., 0.)  (core2)  {Coreset\\Selection};
		\node [text width=0.35cm]   at (7.5,0.)  (dots)   {$\mathbf{\cdots}$};
		\node [box]                at (9., 0.)  (core3)  {Coreset\\Selection};
		\node [start, fill=red!20] at (12., 0.) (finish) {\scriptsize{Finish}};
		
		\path[->, thick] (start) edge node[above] {$T_{\rm warm}$ epochs of} node[below] {\textit{full} training} (core1);
		\path[->, thick] (core1) edge node[above] {$T$ epochs of} node[below] {\textit{subset} training} (core2);
		\path[- , thick] (core2) edge (dots);
		\path[->, thick] (dots.east)  edge (core3);
		\path[->, thick] (core3) edge node[above] {$T$ epochs of} node[below] {\textit{subset} training} (finish);
		\end{tikzpicture}
		\caption{}
	\end{subfigure}\\
	\begin{subfigure}{0.44\textwidth}
		\centering
		\tikzstyle{box} = [rectangle, draw, fill=blue!20, text width=1.5cm, text badly centered, node distance=1cm, minimum height=.85cm, inner sep=0pt, rotate=0]
		\tikzstyle{start} = [circle, draw, fill=green!20, text width=0.6cm, text centered, rounded corners, minimum height=0.6cm]
		\begin{tikzpicture}[node distance = 1cm, thick, scale=0.9, every node/.style={scale=0.75}]
		\draw[thick, fill=blue!20] (-0.90, -0.55) rectangle (2.9, 0.55);
		\node [box, fill=orange!90] at (0., 0.) (core1) {Gradient\\Comput.};
		\node [box, fill=yellow!60] at (2., 0.) (core2) {Greedy\\Selection};
		\path[-> , thick] (core1) edge (core2);
		\end{tikzpicture}
		\caption{\label{fig:coresets_idea_van}}
	\end{subfigure}
	\begin{subfigure}{0.48\textwidth}
		\centering
		\tikzstyle{box} = [rectangle, draw, fill=blue!20, text width=1.5cm, text badly centered, node distance=1cm, minimum height=.85cm, inner sep=0pt, rotate=0]
		\tikzstyle{start} = [circle, draw, fill=green!20, text width=0.6cm, text centered, rounded corners, minimum height=0.6cm]
		\begin{tikzpicture}[node distance = 1cm, thick, scale=0.9, every node/.style={scale=0.75}]
		\draw[thick, fill=blue!20] (-0.90, -0.55) rectangle (4.9, 0.55);
		\node [box, fill=red!60]    at (0., 0.)  (core1) {Advers.\\Attack};
		\node [box, fill=orange!90] at (2., 0.)  (core2) {Gradient\\Comput.};
		\node [box, fill=yellow!60] at (4., 0.)  (core3) {Greedy\\Selection};
		\path[-> , thick] (core1) edge (core2);
		\path[-> , thick] (core2) edge (core3);
		\end{tikzpicture}
		\caption{\label{fig:coresets_idea_adv}}
	\end{subfigure}
	\caption{\label{fig:coresets_idea} Overview of neural network training using coreset selection. (a) Selection is done every $T$ epochs. During the next episodes, the network is only trained on this subset. (b) Coreset selection module for vanilla training. (c) Coreset selection module for adversarial training.}
\end{figure*}
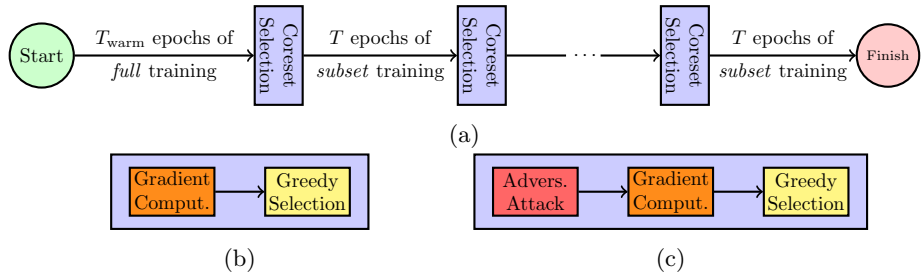

In this paper, we show that the answer to this question is affirmative:
by selecting a \textit{weighted} subset of the data based on the neural network state, we run \textit{weighted} adversarial training only on this selected subset.
We draw an elegant connection between adversarial training and adaptive coreset selection algorithms to achieve this goal.
In particular, we use Danskin's theorem and demonstrate how the entire training data can effectively be approximated with an informative weighted subset.
To conduct this selection, our study shows that one needs to build adversarial examples for the entire training data and solve a respective subset selection objective.
Afterward, training can be performed on this selected subset of the training data.
In our approach, shown in \Cref{fig:coresets_idea}, adversarial coreset selection is only required every few epochs, effectively reducing the training time of robust learning algorithms.
We demonstrate how our proposed method can be used as a general framework in conjunction with different adversarial training objectives, opening the door to a more principled approach for efficient training of robust neural networks in a general setting.
Our experimental results show that one can reduce the training time of various robust training objectives by 2-3 times without sacrificing too much clean or robust accuracy.
In summary, we make the following contributions:
\begin{itemize}\setlength\itemsep{0.25em}
	\item We propose a practical yet principled algorithm for efficient training of robust neural networks based on adaptive coreset selection. To the best of our knowledge, we are the first to use coreset selection in adversarial training.
	\item We show that our approach can be applied to a variety of robust learning objectives, including TRADES~\cite{zhang2019trades}, $\ell_p$-PGD~\cite{madry2018towards} and Perceptual~\cite{laidlaw2021pat} Adversarial Training. Our approach encompasses a broader range of robust models compared to the limited scope of the existing methods. 
	\item Through extensive experiments, we show that the proposed approach can result in a 2-3 fold reduction of the training time, with only a slight reduction in the clean and robust accuracy.
\end{itemize}
\section{Background and Related Work}\label{sec:background}

\subsection{Adversarial Training}\label{sec:sec:adversarial_training}

Let $\mathcal{D}=\left\{\left(\boldsymbol{x}_{i}, y_{i}\right)\right\}_{i=1}^{n} \subset  \mathbb{X} \times \mathbb{C}$ denote a training dataset consisting of $n$ i.i.d.~samples.
Each data point contains an input data $\boldsymbol{x}_{i}$ from domain $\mathbb{X}$ and an associated label $y_{i}$ taking one of $k$ possible values $\mathbb{C}=\left[k\right]=\left\{1, 2, \dots, k\right\}$.
Without loss of generality, in this paper we focus on the image domain $\mathbb{X}$.
Furthermore, assume that $f_{\boldsymbol{\theta}}: \mathbb{X} \rightarrow \mathbb{R}^{k}$ denotes a neural network classifier with parameters $\boldsymbol{\theta}$ that takes $\boldsymbol{x} \in \mathbb{X}$ as input and maps it to a logit value $f_{\boldsymbol{\theta}}(\boldsymbol{x}) \in \mathbb{R}^{k}$.
Then, training a neural network in its most general format can be written as the following minimization problem:
\begin{equation}\label{eq:nn_training_objective}
    \min_{\boldsymbol{\theta}} \sum_{i \in V} \boldsymbol{\Phi} \left(\boldsymbol{x}_{i}, y_{i}; f_{\boldsymbol{\theta}}\right),
\end{equation}
Here, $\boldsymbol{\Phi} \left(\boldsymbol{x}, y; f_{\boldsymbol{\theta}}\right)$ is a function that takes a data point $\left(\boldsymbol{x}, y\right)$ and a function $f_{\boldsymbol{\theta}}$ as its inputs, and its output is a measure of discrepancy between the input $\boldsymbol{x}$ and its ground-truth label~$y$.
Also, $V=\left[n\right]=\left\{1, 2, \dots, n\right\}$ denotes the entire training data.
By writing the training objective in this format, we can denote both vanilla and adversarial training using the same notation.
Below we show how various choices of the function $\boldsymbol{\Phi}$ amount to different training objectives.

\subsubsection{Vanilla Training.}
In case of vanilla training, the function $\boldsymbol{\Phi}$ is a simple evaluation of an appropriate loss function over the neural network output $f_{\boldsymbol{\theta}}(\boldsymbol{x})$ and the ground-truth label $y$.
For instance, for vanilla training we can have:
\begin{equation}\label{eq:vanilla_functional}
    \boldsymbol{\Phi} \left(\boldsymbol{x}, y; f_{\boldsymbol{\theta}}\right) = \mathcal{L}_{\mathrm{CE}}\left(f_{\boldsymbol{\theta}}(\boldsymbol{x}), y\right),
\end{equation}
where $\mathcal{L}_{\mathrm{CE}}(\cdot, \cdot)$ is the cross-entropy loss.

\subsubsection{FGSM, $\ell_p$-PGD, and Perceptual Adversarial Training.}
In these cases, the training objective is itself an optimization problem:
\begin{equation}\label{eq:at_functional}
\boldsymbol{\Phi} \left(\boldsymbol{x}, y; f_{\boldsymbol{\theta}}\right) = \max_{\tilde{\boldsymbol{x}}} \mathcal{L}_{\mathrm{CE}}\left(f_{\boldsymbol{\theta}}(\tilde{\boldsymbol{x}}), y\right)~ \text{s.t.}~\mathrm{d}\left({\tilde{\boldsymbol{x}}, \boldsymbol{x}}\right)\leq \varepsilon
\end{equation}
where $\mathrm{d}(\cdot, \cdot)$~is an appropriate distance measure over image domain~$\mathbb{X}$, and $\varepsilon$ denotes a scalar.
The constraint over $\mathrm{d}({\tilde{\boldsymbol{x}}, \boldsymbol{x}})$ is used to ensure visual similarity between $\tilde{\boldsymbol{x}}$ and $\boldsymbol{x}$.
Solving \Cref{eq:at_functional} amounts to finding an adversarial example $\tilde{\boldsymbol{x}}$ for the clean sample $\boldsymbol{x}$~\cite{madry2018towards}.
Different choices of the visual similarity measure $\mathrm{d}(\cdot, \cdot)$ and solvers for \Cref{eq:at_functional} result in different adversarial training objectives.
\begin{itemize}
	\item FGSM~\cite{goodfellow2014explaining} assumes that ${\mathrm{d}({\tilde{\boldsymbol{x}}, \boldsymbol{x}}) = \norm{{\tilde{\boldsymbol{x}}-\boldsymbol{x}}}_{\infty}}$.
	Using this $\ell_\infty$ assumption, the solution to \Cref{eq:at_functional} is computed using one iteration of gradient ascent.
	\item $\ell_p$-PGD~\cite{madry2018towards} utilizes $\ell_p$ norms as a proxy for visual similarity $\mathrm{d}(\cdot, \cdot)$. Then, several steps of projected gradient ascent is taken to solve \Cref{eq:at_functional}.
	\item Perceptual Adversarial Training~(PAT)~\cite{laidlaw2021pat} replaces  $\mathrm{d}(\cdot, \cdot)$ with \textit{Learned Perceptual Image Patch Similarity}~(LPIPS) distance~\cite{zhang2018lpips}.
	Then, Laidlaw~\etal~\cite{laidlaw2021pat} propose to solve this maximization objective using either projected gradient ascent or Lagrangian relaxation.
\end{itemize}

\subsubsection{TRADES Adversarial Training.}
This approach uses a combination of \Cref{eq:at_functional,eq:vanilla_functional}.
The intuition behind TRADES~\cite{zhang2019trades} is to create a trade-off between clean and robust accuracy.
In particular, the objective is written as:
\begin{align}\label{eq:trades_functional}
\boldsymbol{\Phi} \left(\boldsymbol{x}, y; f_{\boldsymbol{\theta}}\right) =~ \mathcal{L}_{\mathrm{CE}}\left(f_{\boldsymbol{\theta}}(\boldsymbol{x}), y\right)
+ \max_{\tilde{\boldsymbol{x}}} \mathcal{L}_{\mathrm{CE}}\left(f_{\boldsymbol{\theta}}(\tilde{\boldsymbol{x}}), f_{\boldsymbol{\theta}}(\boldsymbol{x})\right)/\lambda, 
\end{align}
such that $\mathrm{d}({\tilde{\boldsymbol{x}}, \boldsymbol{x}})\leq \varepsilon$.
Here, $\lambda$ is a coefficient that controls the trade-off.

\subsection{Coreset Selection}\label{sec:sec:coreset_selection}
Adaptive data subset selection, and \textit{coreset selection} in general, is concerned with finding a weighted subset of the data that can approximate specific attributes of the entire population~\cite{feldman2011coresets}.
Traditionally, coreset selection has been used for different machine learning tasks such as $k$-means and $k$-medians~\cite{harpeled2004oncoresets}, Na\"{i}ve Bayes and nearest neighbor classifiers~\cite{wei2015nnc}, and Bayesian inference~\cite{campbell2018coreset}.

Recently, coreset selection algorithms are being developed for neural network training~\cite{mirzasoleiman2020craig,mirzasoleiman2020crust,killamsetty2021glister,killamsetty2021gradmatch}.
The main idea behind such methods is to approximate the full gradient using a weighted subset of the training data.
These algorithms start with computing the gradient of the loss function with respect to the neural network weights.
This gradient is computed for \textit{every} data sample in the training set.
Then, a selection criterion is formed.
This criterion aims to find a \textit{weighted subset} of the training data that can approximate the full gradient.
In \Cref{sec:proposed_method} we provide a detailed account of these methods.

Existing coreset selection algorithms can only be used for the vanilla training of neural networks.
As such, they still suffer from adversarial vulnerability.
This paper extends coreset selection algorithms to robust neural network training and shows how they can be adopted to various robust training objectives.

\section{Proposed Method}\label{sec:proposed_method}

As discussed in \Cref{sec:introduction}, the main bottleneck in the time/computational complexity of adversarial training stems from constructing adversarial examples for the entire training set at each epoch.
FAT~\cite{wong2020fast} tries to eliminate this issue by using FGSM as its adversarial example generator.
However, this simplification 1) may lead to catastrophic overfitting~\cite{wong2020fast,andriushchenko2020understanding}, and 2) is not easy to generalize to all types of adversarial training as FGSM is designed explicitly for $\ell_\infty$ attacks.

Instead of using a faster adversarial example generator, here we take a different, \textit{orthogonal} path and try to reduce the training set size effectively.
This way, the original adversarial training algorithm can still be used on this smaller subset of training data. 
This approach can reduce the training time while optimizing a similar objective as the original training.
In this sense, it leads to a more \textit{unified} method that can be used along with various types of adversarial training objectives, including the ones that already exist and the ones that will be proposed in the future. 

The main hurdle in materializing this idea is the following question: \textit{How should we select this subset of the training data without hurting either the clean or robust accuracy?}
To answer this question, we propose to use coreset selection on the training data to reduce the sample size and improve training efficiency.

\subsection{Problem Statement}\label{sec:sec:coreset_statement}

Using our general notation from \Cref{sec:sec:adversarial_training}, we write both vanilla and adversarial training using the same objective:
\begin{equation}\label{eq:nn_training_objective_repeat}
\min_{\boldsymbol{\theta}} \sum_{i \in \textcolor{red}{V}} \boldsymbol{\Phi} \left(\boldsymbol{x}_{i}, y_{i}; f_{\boldsymbol{\theta}}\right),
\end{equation}
where $\textcolor{red}{V}$ denotes \textcolor{red}{the entire training data}, and depending on the training task, $\boldsymbol{\Phi} \left(\boldsymbol{x}_{i}, y_{i}; f_{\boldsymbol{\theta}}\right)$ takes any of the \Cref{eq:vanilla_functional,eq:at_functional,eq:trades_functional} forms.
We adopt this notation to make our analysis more accessible.

As discussed in \Cref{sec:sec:coreset_selection}, coreset selection can be seen as a two-step process.
First, the gradient of the loss function with respect to the neural network weights is computed for each training sample.
Then, based on the gradients obtained in step one, a weighted subset~(a.k.a.~the coreset) of the training data is formed~(see \Cref{fig:coresets_idea_van}).
This subset is obtained such that the weighted gradients of the samples inside the coreset can provide a good approximation of the full gradient.

Specifically, using our universal notation in \Cref{eq:nn_training_objective_repeat}, we write coreset selection for both vanilla and adversarial training as:
\begin{equation}\label{eq:gradient_app_coreset}
	\min_{\textcolor{mygreen}{S} \subseteq \textcolor{red}{V}, \textcolor{mygreen}{\boldsymbol{\gamma}}} \norm{\sum_{i \in \textcolor{red}{V}} \nabla_{\boldsymbol{\theta}}\boldsymbol{\Phi} \left(\boldsymbol{x}_{i}, y_{i}; f_{\boldsymbol{\theta}}\right)-\sum_{j \in \textcolor{mygreen}{S}} \textcolor{mygreen}{\gamma}_{j}\nabla_{\boldsymbol{\theta}}\boldsymbol{\Phi} \left(\boldsymbol{x}_{j}, y_{j}; f_{\boldsymbol{\theta}}\right)},
\end{equation}
where $\textcolor{mygreen}{S} \subseteq \textcolor{red}{V}$ is \textcolor{mygreen}{the coreset}, and $\textcolor{mygreen}{\gamma}_{j}$'s are \textcolor{mygreen}{the weights} of each sample in the coreset.
Once the coreset $\textcolor{mygreen}{S}$ is found, instead of training the neural network using \Cref{eq:nn_training_objective_repeat}, we can optimize its parameters using a weighted objective over the coreset:
\begin{equation}\label{eq:nn_training_objective_coreset}
    \min_{\boldsymbol{\theta}} \sum_{j \in \textcolor{mygreen}{S}} \textcolor{mygreen}{\gamma}_{j} \boldsymbol{\Phi} \left(\boldsymbol{x}_{j}, y_{j}; f_{\boldsymbol{\theta}}\right).
\end{equation}

It can be shown that solving \Cref{eq:gradient_app_coreset} is NP-hard~\cite{mirzasoleiman2020craig,mirzasoleiman2020crust}.
Roughly, various coreset selection methods differ in how they approximate the solution of the aforementioned objective.
For instance, \textsc{Craig}~\cite{mirzasoleiman2020craig} casts this objective as a \textit{submodular set cover problem} and uses existing greedy solvers to get an approximate solution.
As another example, \textsc{GradMatch}~\cite{killamsetty2021gradmatch} analyzes the convergence of stochastic gradient descent using adaptive data subset selection.
Based on this study, Killamsetty~\etal~\cite{killamsetty2021gradmatch} propose to use Orthogonal Matching Pursuit~(OMP)~\cite{pati1992omp,elenberg2016restricted} as a greedy solver of the data selection objective.
More information about these methods is provided in \Cref{ap:sec:greedy_selection}.

The issue with the aforementioned coreset selection methods is that they are designed explicitly for vanilla training of neural networks~(see \Cref{fig:coresets_idea_van}), and they do not reflect the requirements of adversarial training.
As such, we should modify these methods to make them suitable for our purpose of robust neural network training.
Meanwhile, we should also consider the fact that the field of coreset selection is still evolving.
Thus, we aim to find a general modification that can later be used alongside newer versions of greedy coreset selection algorithms.

We notice that various coreset selection methods proposed for vanilla neural network training only differ in their choice of greedy solvers.
Therefore, we narrow down the changes we want to make to the first step of coreset selection: gradient computation.
Then, existing greedy solvers can be used to find the subset of training data that we are looking for.
To this end, we draw a connection between coreset selection methods and adversarial training using Danskin's theorem, as outlined next.
Our analysis shows that for adversarial coreset selection, one needs to add a pre-processing step where adversarial attacks for the raw training data need to be computed~(see~\Cref{fig:coresets_idea_adv}).

\subsection{Coreset Selection for Efficient Adversarial Training}\label{sec:sec:coreset_adversarial}

As discussed above, to construct the \Cref{eq:gradient_app_coreset} objective, we need to compute the loss gradient with respect to the neural network weights.
Once done, we can use existing greedy solvers to find the solution.
The gradient computation needs to be performed for the entire training set.
In particular, using our notation from \Cref{sec:sec:adversarial_training}, this step can be written as:
\begin{equation}\label{eq:nn_gradient}
\nabla_{\boldsymbol{\theta}}\boldsymbol{\Phi} \left(\boldsymbol{x}_{i}, y_{i}; f_{\boldsymbol{\theta}}\right) \quad \forall\quad i \in V,	
\end{equation}
where $V$ denotes the training set.

For vanilla neural network training~(see \Cref{sec:sec:adversarial_training}) the above gradient is simply equal to $\nabla_{\boldsymbol{\theta}}\mathcal{L}_{\mathrm{CE}}\left(f_{\boldsymbol{\theta}}(\boldsymbol{x}_{i}), y_{i}\right)$ which can be computed using standard backpropagation.
In contrast, for the adversarial training objectives in \Cref{eq:at_functional,eq:trades_functional}, this gradient requires taking partial derivative of a maximization objective.
To this end, we use the famous Dasnkin's theorem~\cite{danskin1967theory} as stated below.

\begin{theorem}[Theorem A.1 \cite{madry2018towards}]\label{danskin_theorem}
	Let $\mathcal{S}$ be a nonempty compact topological space, $\ell: \mathbb{R}^{m} \times \mathcal{S} \rightarrow \mathbb{R}$ be such that $\ell(\cdot, \boldsymbol{\delta})$ is differentiable for every $\boldsymbol{\delta} \in \mathcal{S}$, and $\nabla_{\boldsymbol{\theta}} \ell(\boldsymbol{\theta}, \boldsymbol{\delta})$ is continuous on $\mathbb{R}^{m} \times \mathcal{S}$.
	Also, let ${\boldsymbol{\delta}^{*}(\boldsymbol{\theta})=\left\{\boldsymbol{\delta} \in \arg \max _{\boldsymbol{\delta} \in \mathcal{S}} \ell(\boldsymbol{\theta}, \boldsymbol{\delta})\right\}}$.
	Then, the corresponding max-function $\phi(\boldsymbol{\theta})=\max _{\delta \in \mathcal{S}} \ell(\boldsymbol{\theta}, \boldsymbol{\delta})$ is locally Lipschitz continuous, directionally differentiable, and its directional derivatives along vector $\boldsymbol{h}$ satisfy:
	$$
	\phi^{\prime}(\boldsymbol{\theta}, \boldsymbol{h})=\sup _{\boldsymbol{\delta} \in \boldsymbol{\delta}^{*}(\boldsymbol{\theta})} \boldsymbol{h}^{\top} \nabla_{\boldsymbol{\theta}} \ell(\boldsymbol{\theta}, \boldsymbol{\delta}).
	$$
	In particular, if for some $\boldsymbol{\theta} \in \mathbb{R}^{m}$ the set $\boldsymbol{\delta}^{*}(\boldsymbol{\theta})=\left\{\boldsymbol{\delta}_{\boldsymbol{\theta}}^{*}\right\}$ is a singleton, then the max-function is differentiable at $\boldsymbol{\theta}$ and
	$$
	\nabla \phi(\boldsymbol{\theta})=\nabla_{\boldsymbol{\theta}} \ell\left(\boldsymbol{\theta}, \boldsymbol{\delta}_{\boldsymbol{\theta}}^{*}\right).
	$$
\end{theorem}

In summary, \Cref{danskin_theorem} indicates how to take the gradient of a max-function.
To this end, it suffices to 1) find the maximizer, and 2) evaluate the normal gradient at this point.

Now that we have stated Danskin's theorem, we are ready to show how it can provide the connection between coreset selection and the adversarial training objectives of \Cref{eq:at_functional,eq:trades_functional}.
We do this for the two cases of adversarial training and TRADES as outlined next.

\subsubsection{\textcolor{blue}{Case 1.} ($\ell_p$-PGD and Perceptual Adversarial Training)}
Going back to \Cref{eq:nn_gradient}, we know that to perform coreset selection, we need to compute this gradient term for our objective in \Cref{eq:at_functional}.
In other words, we need to compute:
\begin{equation}\label{eq:nn_gradient_at}
\nabla_{\boldsymbol{\theta}}\boldsymbol{\Phi} \left(\boldsymbol{x}, y; f_{\boldsymbol{\theta}}\right) =  \nabla_{\boldsymbol{\theta}} \max_{\tilde{\boldsymbol{x}}} \mathcal{L}_{\mathrm{CE}}\left(f_{\boldsymbol{\theta}}(\tilde{\boldsymbol{x}}), y\right)
\end{equation}
under the constraint $\mathrm{d}\left({\tilde{\boldsymbol{x}}, \boldsymbol{x}}\right)\leq \varepsilon$ for every training sample.
Based on Danskin's theorem, we can deduce:
\begin{equation}\label{eq:nn_gradient_at_penufinal}
    \nabla_{\boldsymbol{\theta}}\boldsymbol{\Phi} \left(\boldsymbol{x}, y; f_{\boldsymbol{\theta}}\right) =  \nabla_{\boldsymbol{\theta}} \mathcal{L}_{\mathrm{CE}}\left(f_{\boldsymbol{\theta}}({\boldsymbol{x}^{*}}), y\right),
\end{equation}
where $\boldsymbol{x}^{*}$ is the solution to:
\begin{equation}\label{eq:at_max_objective}
    \argmax_{\tilde{\boldsymbol{x}}} \mathcal{L}_{\mathrm{CE}}\left(f_{\boldsymbol{\theta}}(\tilde{\boldsymbol{x}}), y\right) \quad \text{s.t.} \quad \mathrm{d}\left({\tilde{\boldsymbol{x}}, \boldsymbol{x}}\right)\leq \varepsilon.
\end{equation}
The conditions under which Danskin's theorem hold might not be satisfied for neural networks in general.
This is due to the presence of functions with discontinuous gradients, such as ReLU activation, in neural networks.
More importantly, finding the exact solution of \Cref{eq:at_max_objective} is not straightforward as neural networks are highly non-convex.
Usually, the exact solution $\boldsymbol{x}^{*}$ is replaced with its approximation, which is an adversarial example generated under the \Cref{eq:at_max_objective} objective~\cite{madry2018adversarial}.
Based on this approximation, we can re-write \Cref{eq:nn_gradient_at_penufinal} as:
\begin{equation}\label{eq:nn_gradient_at_final}
    \nabla_{\boldsymbol{\theta}}\boldsymbol{\Phi} \left(\boldsymbol{x}, y; f_{\boldsymbol{\theta}}\right) \approx  \nabla_{\boldsymbol{\theta}} \mathcal{L}_{\mathrm{CE}}\left(f_{\boldsymbol{\theta}}({\boldsymbol{x}_{\rm{adv}}}), y\right).
\end{equation}
In other words, to perform coreset selection for $\ell_p$-PGD~\cite{madry2018towards} and Perceptual~\cite{laidlaw2021pat} Adversarial Training, one needs to add a pre-processing step to the gradient computation.
At this step, adversarial examples for the entire training set must be constructed.
Then, the coresets can be built as in vanilla neural networks.

\subsubsection{\textcolor{blue}{Case 2.} (TRADES Adversarial Training)}

For TRADES~\cite{zhang2019trades}, the gradient computation is slightly different as the objective in \Cref{eq:trades_functional} consists of two terms.
In this case, the gradient can be written as:
\begin{align}\label{eq:nn_gradeint_trades}
\nabla_{\boldsymbol{\theta}}\boldsymbol{\Phi} \left(\boldsymbol{x}, y; f_{\boldsymbol{\theta}}\right) = \nabla_{\boldsymbol{\theta}}\mathcal{L}_{\mathrm{CE}}\left(f_{\boldsymbol{\theta}}(\boldsymbol{x}), y\right) + \nabla_{\boldsymbol{\theta}}\max_{\tilde{\boldsymbol{x}}} \mathcal{L}_{\mathrm{CE}}\left(f_{\boldsymbol{\theta}}(\tilde{\boldsymbol{x}}), f_{\boldsymbol{\theta}}(\boldsymbol{x})\right)/\lambda,
\end{align}
where $\mathrm{d}({\tilde{\boldsymbol{x}}, \boldsymbol{x}})\leq \varepsilon$.
The first term is the normal gradient of the neural network.
For the second term, we apply Danskin's theorem to obtain:
\begin{align}\label{eq:nn_gradeint_trades_penufinal}
    \nabla_{\boldsymbol{\theta}}\boldsymbol{\Phi} \left(\boldsymbol{x}, y; f_{\boldsymbol{\theta}}\right) \approx \nabla_{\boldsymbol{\theta}}\mathcal{L}_{\mathrm{CE}}\left(f_{\boldsymbol{\theta}}(\boldsymbol{x}), y\right) + \nabla_{\boldsymbol{\theta}} \mathcal{L}_{\mathrm{CE}}\left(f_{\boldsymbol{\theta}}(\boldsymbol{x}_{\rm{adv}}), f_{\boldsymbol{\theta}}(\boldsymbol{x})\right)/\lambda,
\end{align}
where $\boldsymbol{x}_{\rm{adv}}$ is an approximate solution to:
\begin{equation}\label{eq:trades_max}
    \argmax_{\tilde{\boldsymbol{x}}} \mathcal{L}_{\mathrm{CE}}\left(f_{\boldsymbol{\theta}}(\tilde{\boldsymbol{x}}), f_{\boldsymbol{\theta}}(\boldsymbol{x})\right)/\lambda \quad \text{s.t.} \quad \mathrm{d}\left({\tilde{\boldsymbol{x}}, \boldsymbol{x}}\right)\leq \varepsilon.
\end{equation}

Having found the loss gradients $\nabla_{\boldsymbol{\theta}}\boldsymbol{\Phi} \left(\boldsymbol{x}_{i}, y_{i}; f_{\boldsymbol{\theta}}\right)$ for $\ell_p$-PGD, PAT~(\textcolor{blue}{Case 1}), and TRADES~(\textcolor{blue}{Case 2}), we can construct \Cref{eq:gradient_app_coreset} and use existing greedy solvers like \textsc{Craig}~\cite{mirzasoleiman2020craig} or \textsc{GradMatch}~\cite{killamsetty2021gradmatch} to find the coreset.
As we saw, adversarial coreset selection requires adding a pre-processing step where we need to build perturbed versions of the training data using their respective objectives in \Cref{eq:at_max_objective,eq:trades_max}.
Then, the gradients are computed using \Cref{eq:nn_gradient_at_final,eq:nn_gradeint_trades_penufinal}.
Afterward, greedy subset selection algorithms are used to construct the coresets based on the value of the gradients.
Finally, having selected the coreset data, one can run \textit{weighted} adversarial training only on the data that remains in the coreset.
As can be seen, we are not changing the essence of the training objective in this process.
We are just reducing the dataset size to enhance our proposed solution's computational efficiency; as such, we can use it along with any adversarial training objective.

\subsection{Practical Considerations}\label{sub:sub:practical}

Since coreset selection depends on the current values of the neural network weights, it is important to update the coresets as the training evolves.
Prior work~\cite{killamsetty2021glister,killamsetty2021gradmatch} has shown that this selection needs to be done every $T$ epochs, where $T$ is usually greater than 15.
Also, we employ small yet crucial practical changes while using coreset selection to increase efficiency.
We summarize these practical tweaks below.
Further detail can be found in~\cite{killamsetty2021gradmatch,mirzasoleiman2020craig}.
\paragraph{Gradient Approximation.}
	As we saw, both \Cref{eq:nn_gradient_at_final,eq:nn_gradeint_trades_penufinal} require computation of the loss gradient with respect to the neural network weights.
	This is equal to backpropagation through the entire neural network, which is not very efficient.
	Instead, it is common to replace the exact gradients in \Cref{eq:nn_gradient_at_final,eq:nn_gradeint_trades_penufinal} with their last-layer approximation~\cite{katharopoulos2018notall,mirzasoleiman2020craig,killamsetty2021gradmatch}.
	In other words, instead of backpropagating through the entire network, one can backpropagate up until the penultimate layer.
	This estimate has an approximate complexity equal to forwardpropagation, and it has been shown to work well in practice~\cite{mirzasoleiman2020craig,mirzasoleiman2020crust,killamsetty2021glister,killamsetty2021gradmatch}.
\paragraph{Batch-wise Coreset Selection.}
	As discussed in \Cref{sec:sec:coreset_adversarial}, data selection is usually done in a \textit{sample-wise} fashion where each data sample is separately considered to be selected.
	This way, one must find the data candidates from the entire training set.
	To increase efficiency, Killamsetty~\etal~\cite{killamsetty2021gradmatch} proposed the \textit{batch-wise} variant.
	In this type of coreset selection, the data is first split into several batches.
	Then, the algorithm makes a selection out of these batches.
	Intuitively, this change increases efficiency as the sample size is reduced from the number of data points to the number of batches.
\paragraph{Warm-start with the Entire Data.}
	Finally, we warm-start the training using the entire dataset.
	Afterward, coreset selection is activated, and training is only performed using the data in the coreset.

\subsubsection{Final Algorithm}
\Cref{fig:coresets_idea} and \Cref{alg:adv_core} in \Cref{ap:final_alg} summarize our coreset selection approach for adversarial training.
As can be seen, our proposed method is a generic and principled approach in contrast to existing methods such as FAT~\cite{wong2020fast}.
In particular, our approach provides the following advantages compared to existing methods:
\begin{enumerate}
	\item The proposed approach does not involve algorithmic level manipulations and dependency on specific training attributes such as $\ell_\infty$ bound or cyclic learning rate.
	Also, it controls the training speed through coreset size, which can be specified solely based on available computational resources.
	\item The simplicity of our method makes it compatible with any existing/future adversarial training objectives. Furthermore, as we will see in \Cref{sec:experiments}, our approach can be combined with any greedy coreset selection algorithms to deliver robust neural networks.
\end{enumerate}
These characteristics increase the likelihood of applying our proposed method for robust neural network training no matter the training objective.
This contrasts with existing methods that solely focus on a particular training objective.

\section{Experimental Results}\label{sec:experiments}

In this section, we present our experimental results.\footnote{Our implementation can be found in \href{https://github.com/hmdolatabadi/ACS}{this repository}.}
We show how our proposed approach can efficiently reduce the training time of various robust objectives in different settings.
To this end, we train neural networks using TRADES~\cite{zhang2019trades}, $\ell_p$-PGD~\cite{madry2018towards} and PAT~\cite{laidlaw2021pat} on CIFAR-10~\cite{krizhevsky2009learning}, SVHN~\cite{netzer2011reading}, and a subset of ImageNet~\cite{russakovsky2015imagenet} with 12 classes.
For TRADES and $\ell_p$-PGD training, we use ResNet-18~\cite{he2016deep} classifiers, while for PAT we use ResNet-50 architectures.

\begin{table}[t!]
	\caption{Clean~(ACC) and robust~(RACC) accuracy, and total training time~(T) of different adversarial training methods.
		For each objective, all the hyper-parameters were kept the same as full training.
		For our proposed approach, the difference with full training is shown in parentheses.
		The results are averaged over 5 runs.
		More detail can be found in \Cref{ap:sec:imp_det}.}
	\label{tab:TRADES_lp_PGD}
	\begin{center}
		\begin{scriptsize}
		    \setlength\tabcolsep{0.45em}
			\def\arraystretch{1.5}
			\begin{tabular}{cccccc}
				\toprule
				\multirow{2}{*}{\rotatebox[origin=c]{90}{\tiny{\textbf{Objec.}}}}
				&\multirow{2}{*}{\rotatebox[origin=c]{90}{\tiny{\textbf{Data}}}}
				&\multirow{2}{*}{\textbf{Training Method}}
				&\multicolumn{3}{c}{\textbf{Performance Measures}}\\
				\cmidrule(lr){4-6}
				&&                                 &  \textuparrow~\textbf{ACC} (\%)         & \textuparrow~\textbf{RACC} (\%)         & \textdownarrow~\textbf{T} (mins)\\
				\midrule
				\multirow{3}{*}{\rotatebox[origin=c]{90}{\tiny{\textbf{TRADES}}}}  & \multirow{3}{*}{\rotatebox[origin=c]{90}{\tiny{\textbf{CIFAR-10}}}}
				&  Adv. \textsc{Craig} (Ours)      & $83.03$ (\textcolor{red}{$-2.38$})            & $41.45$ (\textcolor{red}{$-2.74$})            &  $179.20$ (\textcolor{darkgreen}{$-165.09$})\\
				&& Adv. \textsc{GradMatch} (Ours)  & $83.07$ (\textcolor{red}{$-2.34$})            & $41.52$ (\textcolor{red}{$-2.67$})            &  $178.73$ (\textcolor{darkgreen}{$-165.56$})\\
				&& Full Adv. Training              & $85.41$                                       & $44.19$                                       &  $344.29$\\
				\midrule
				\multirow{3}{*}{\rotatebox[origin=c]{90}{\tiny{\textbf{$\ell_\infty$-PGD}}}} & \multirow{3}{*}{\rotatebox[origin=c]{90}{\tiny{\textbf{CIFAR-10}}}}
				&  Adv. \textsc{Craig} (Ours)      & $80.37$ (\textcolor{red}{$-2.77$})            & $45.07$ (\textcolor{darkgreen}{$+3.68$})            &  $148.01$ (\textcolor{darkgreen}{$-144.86$})\\
				&& Adv. \textsc{GradMatch} (Ours)  & $80.67$ (\textcolor{red}{$-2.47$})            & $45.23$ (\textcolor{darkgreen}{$+3.84$})            &  $148.03$ (\textcolor{darkgreen}{$-144.84$})\\
				&& Full Adv. Training              & $83.14$                                       & $41.39$                                             &  $292.87$\\
				\midrule
				\multirow{3}{*}{\rotatebox[origin=c]{90}{\tiny{\textbf{$\ell_2$-PGD}}}} & \multirow{3}{*}{\rotatebox[origin=c]{90}{\tiny{\textbf{SVHN}}}}
				&  Adv. \textsc{Craig} (Ours)      & $95.42$ (\textcolor{darkgreen}{$+0.10$})            & $49.68$ (\textcolor{red}{$-3.34$})            &  $130.04$ (\textcolor{darkgreen}{$-259.42$})\\
				&& Adv. \textsc{GradMatch} (Ours)  & $95.57$ (\textcolor{darkgreen}{$+0.25$})            & $50.41$ (\textcolor{red}{$-2.61$})            &  $125.53$ (\textcolor{darkgreen}{$-263.93$})\\
				&& Full Adv. Training              & $95.32$                                             & $53.02$                                       &  $389.46$\\
				\bottomrule
			\end{tabular}
		\end{scriptsize}
	\end{center}
\end{table}

\subsection{TRADES and $\ell_p$-PGD Robust Training}
In our first experiments, we train ResNet-18 classifiers on CIFAR-10 and SVHN datasets using TRADES, $\ell_\infty$ and $\ell_2$-PGD adversarial training objectives.
In each case, we set the training hyper-parameters such as the learning rate, the number of epochs, and attack parameters.
Then, we train the network using the entire training data and our adversarial coreset selection approach.
For our approach, we use batch-wise versions of \textsc{Craig}~\cite{mirzasoleiman2020craig} and \textsc{GradMatch}~\cite{killamsetty2021gradmatch} with warm-start.
We set the \textit{coreset size} (the percentage of training data to be selected) to \textit{50\%} for CIFAR-10 and \textit{30\%} for SVHN to get a reasonable balance between accuracy and training time.
We report the clean and robust accuracy (in \%) as well as the total training time (in minutes) in \Cref{tab:TRADES_lp_PGD}.
For our approach, we also report the difference with full training in parentheses.
In each case, we evaluate the robust accuracy using an attack with similar attributes as the training objective~(for more information, see \Cref{ap:sec:imp_det}).

As seen, in all cases, we reduce the training time by more than a factor of two while keeping the clean and robust accuracy almost intact.
Note that in these experiments, all the training attributes such as the hyper-parameters, learning rate scheduler, etc. are the same among different training schemes.
This is important since we want to clearly show the relative boost in performance that one can achieve just by using coreset selection.
Nonetheless, it is likely that by tweaking the hyper-parameters of our approach, one can obtain even better results in terms of clean and robust accuracy.

\begin{table}[t!]
	\caption{Clean (ACC) and robust (RACC) accuracy and total training time (T) of Perceptual Adversarial Training for CIFAR-10 and ImageNet-12 datasets.
	         At inference, the networks are evaluated against five attacks that were not seen during training (Unseen RACC) and different versions of Perceptual Adversarial Attack (Seen RACC).
	         In each case, the average is reported.
	         For more information and details about the experiment, please see the \Cref{ap:sec:imp_det,ap:extended}.}
	\label{tab:LPA_sum}
	\begin{center}
		\begin{scriptsize}
			\setlength\tabcolsep{0.3em}
			\def\arraystretch{1.75}
			\begin{tabular}{ccccccc}
				\toprule
				\multirow{2}{*}{\rotatebox[origin=c]{90}{\textbf{Data}}}
				& \multirow{2}{*}{\textbf{Training Method}}
				& \multirow{2}{*}{\textuparrow~\textbf{ACC} (\%)}
				& \multicolumn{2}{c}{\textuparrow~\textbf{RACC} (\%)}
				& \multirow{2}{*}{\textdownarrow~\textbf{T} (mins)}\\
				\cmidrule{4-5}
				& & & Unseen & Seen\\
				\midrule
				\multirow{3}{*}{\rotatebox[origin=c]{90}{\textbf{CIFAR-10}}}
				& Adv. \textsc{Craig} (Ours)             & $83.21$ (\textcolor{red}{$-2.81$})         & $46.55$ (\textcolor{red}{$-1.49$})    & $13.49$ (\textcolor{red}{$-1.83$})    & $767.34$ (\textcolor{darkgreen}{$-915.60$}) \\
				& Adv. \textsc{GradMatch} (Ours)         & $83.14$ (\textcolor{red}{$-2.88$}) 	      & $46.11$ (\textcolor{red}{$-1.93$})    & $13.74$ (\textcolor{red}{$-1.54$})    & $787.26$ (\textcolor{darkgreen}{$-895.68$}) \\
				& Full PAT (Fast-LPA)                    & $86.02$	                                  & $48.04$                               & $15.32$                               & $1682.94$ \\
				\midrule
				\multirow{3}{*}{\rotatebox[origin=c]{90}{\textbf{ImageNet}}}
				& Adv. \textsc{Craig} (Ours)             & $86.99$ (\textcolor{red}{$-4.23$})         & $53.05$ (\textcolor{red}{$-0.18$})    & $22.56$ (\textcolor{red}{$-0.77$})   & $2817.06$ (\textcolor{darkgreen}{$-2796.06$}) \\
				& Adv. \textsc{GradMatch} (Ours)         & $87.08$ (\textcolor{red}{$-4.14$}) 	      & $53.17$ (\textcolor{red}{$-0.06$})    & $20.74$ (\textcolor{red}{$-2.59$})   & $2865.72$ (\textcolor{darkgreen}{$-2747.40$}) \\
				& Full PAT (Fast-LPA)                    & $91.22$	                                  & $53.23$                               & $23.33$                              & $5613.12$ \\
				\bottomrule
			\end{tabular}
		\end{scriptsize}
	\end{center}
\end{table}

\subsection{Perceptual Adversarial Training vs. Unseen Attacks}
As discussed in \Cref{sec:background}, PAT~\cite{laidlaw2021pat} replaces the visual similarity measure~$\mathrm{d}(\cdot, \cdot)$ in \Cref{eq:at_functional} with LPIPS~\cite{zhang2018lpips} distance.
The logic behind this choice is that $\ell_p$ norms can only capture a small portion of images similar to the clean one, limiting the search space of adversarial attacks.
Motivated by this reason, Laidlaw~\etal~\cite{laidlaw2021pat} propose two different ways of finding the solution to \Cref{eq:at_functional} when $\mathrm{d}(\cdot, \cdot)$ is the LPIPS distance.
The first version uses PGD, and the second is a relaxation of the original problem using the Lagrangian form.
We refer to these two versions as PPGD~(Perceptual PGD) and LPA~(Lagrangian Perceptual Attack), respectively.
Then, Laidlaw~\etal~\cite{laidlaw2021pat} proposed to utilize a fast version of LPA to enable its efficient usage in adversarial training.

For our next set of experiments, we show how our approach can be adapted to this unusual training objective.
This is done to showcase the compatibility of our proposed method with different training objectives as opposed to existing methods that are carefully tuned for a particular training objective.
To this end, we train ResNet-50 classifiers using Fast-LPA.
We train the classifiers on CIFAR-10 and ImageNet-12 datasets.
Like our previous experiments, we set the hyper-parameters of the training to be fixed and then train the models using the entire training data and our adversarial coreset selection method.
For our method, we use batch-wise versions of \textsc{Craig}~\cite{mirzasoleiman2020craig} and \textsc{GradMatch}~\cite{killamsetty2021gradmatch} with warm-start.
The \textit{coreset size} for CIFAR-10 and ImageNet-12 were set to \textit{40\%} and \textit{50\%}, respectively.
We measure the performance of the trained models against unseen attacks during training and the two variants of perceptual attacks as in~\cite{laidlaw2021pat}.
The unseen attacks for each dataset were selected similarly to~\cite{laidlaw2021pat}.
We also record the total training time taken by each method.

\Cref{tab:LPA_sum} summarizes our results on PAT using Fast-LPA (full results can be found in \Cref{ap:extended}).
As seen, our adversarial coreset selection approach can deliver a competitive performance in terms of clean and average unseen attack accuracy while reducing the training time by at least a factor of two.
These results indicate the flexibility of our adversarial coreset selection that can be combined with various objectives.
This is due to the orthogonality of the proposed approach with the existing efficient adversarial training methods.
In this case, we see that we can make Fast-LPA even faster using our approach.

\subsection{Compatibility with Existing Methods}
To showcase that our adversarial coreset selection approach is complementary to existing methods, we integrate it with a stable version of Fast Adversarial Training~(FAT)~\cite{wong2020fast} that does not use a cyclic learning rate.
Specifically, we train a neural network using FAT~\cite{wong2020fast}, and then add adversarial coreset selection to this approach and record the training time and clean/robust accuracy.
We run the experiments on the CIFAR-10 dataset and train a ResNet-18 for each case.
We set the \textit{coreset size} to \textit{50\%} for our methods.
The results are shown in \Cref{tab:FAT}.
As can be seen, our approach can be easily combined with existing methods to deliver faster training.
This is due to the orthogonality of our approach that we discussed previously.

\begin{table}[t!]
	\caption{Clean~(ACC) and robust~(RACC) accuracy, and average training speed~(S\textsubscript{avg}) of Fast Adversarial Training~\cite{wong2020fast} without and with our adversarial coreset selection on CIFAR-10.
	         The difference with full training is shown in parentheses for our proposed approach.}
	\label{tab:FAT}
	\begin{center}
		\begin{scriptsize}
		    \setlength\tabcolsep{0.35em}
			\def\arraystretch{1.25}
			\begin{tabular}{lcccc}
				\toprule
				\multirow{2}{*}{\textbf{Training Method}}
				&\multicolumn{4}{c}{\textbf{Performance Measures}}\\
				\cmidrule(lr){2-5}
				&  \textuparrow~\textbf{ACC} (\%)                           & \textuparrow~\textbf{RACC} (\%)                           & \textdownarrow~\textbf{S}\textsubscript{avg} (min/epoch)    & \textdownarrow~\textbf{T} (min)\\
				\midrule
				Fast Adv. Training                & $86.20$                                       & $47.54$                                       &  $0.5178$  &  $31.068$\\
				~+ Adv. \textsc{Craig} (Ours)      & $82.56$ (\textcolor{red}{$-3.64$})            & $47.77$ (\textcolor{darkgreen}{$+0.23$})      &  $0.2783$  &  $16.695$ (\textcolor{darkgreen}{$-14.373$})\\
				~+ Adv. \textsc{GradMatch} (Ours)   & $82.53$ (\textcolor{red}{$-3.67$})            & $47.88$ (\textcolor{darkgreen}{$+0.34$})      &  $0.2737$  &  $16.419$ (\textcolor{darkgreen}{$-14.649$})\\
				\bottomrule
			\end{tabular}
		\end{scriptsize}
	\end{center}
\end{table}
\begin{figure}[tb!]
		\centering
		\begin{minipage}{.38\linewidth}
			    \centering
                \includegraphics[width=1.0\linewidth]{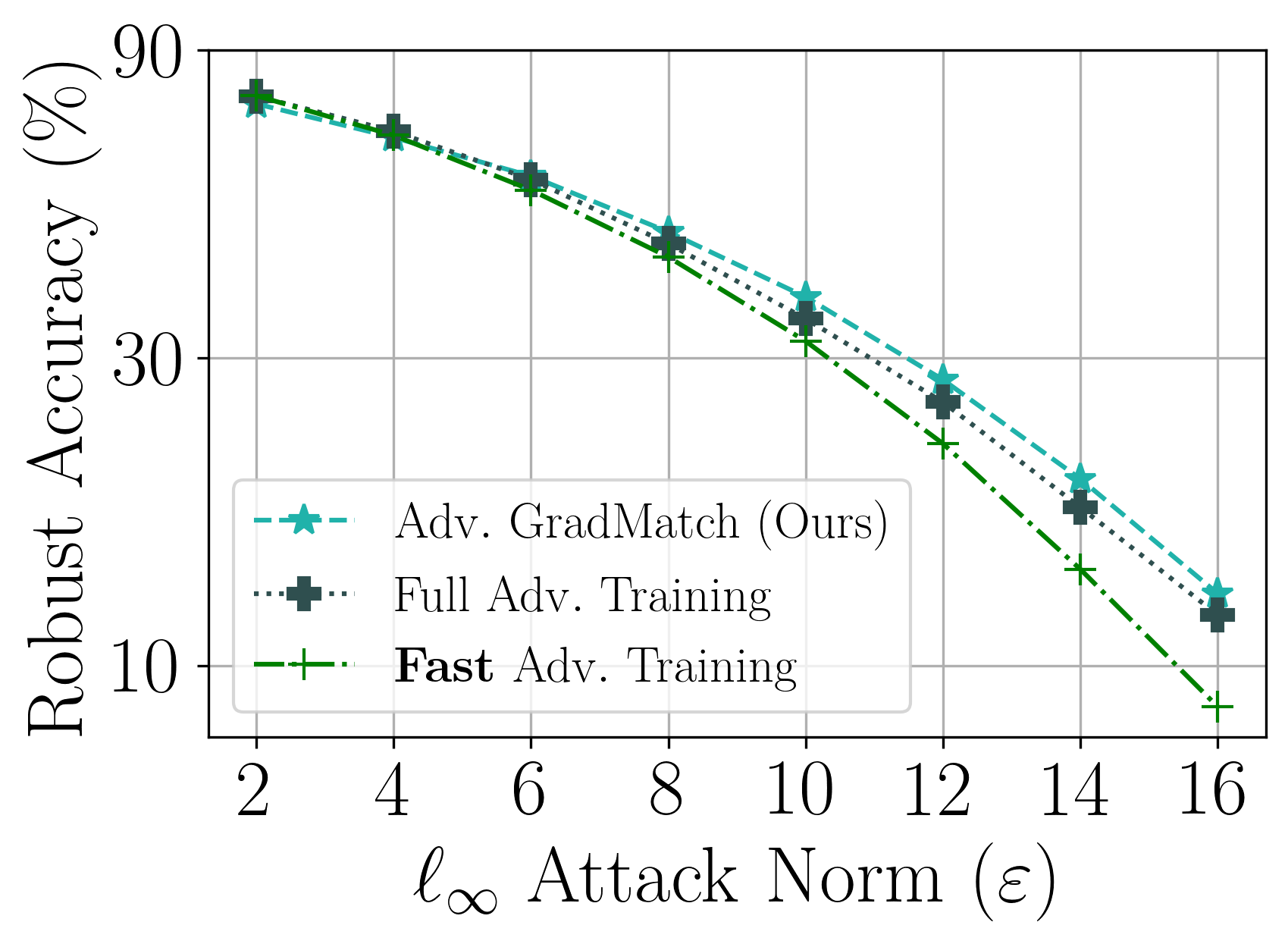}
                \caption{Robust accuracy as a function of $\ell_\infty$ attack norm. We train neural networks with a perturbation norm of $\norm{\varepsilon}_\infty \leq 8$ on CIFAR-10.
                         At inference, we evaluate the robust accuracy against PGD-50 with various attack strengths.}
			\label{fig:fat}
		\end{minipage}
		\hfill
	    \begin{minipage}{.58\linewidth}
			    \centering
                \includegraphics[width=1.0\linewidth]{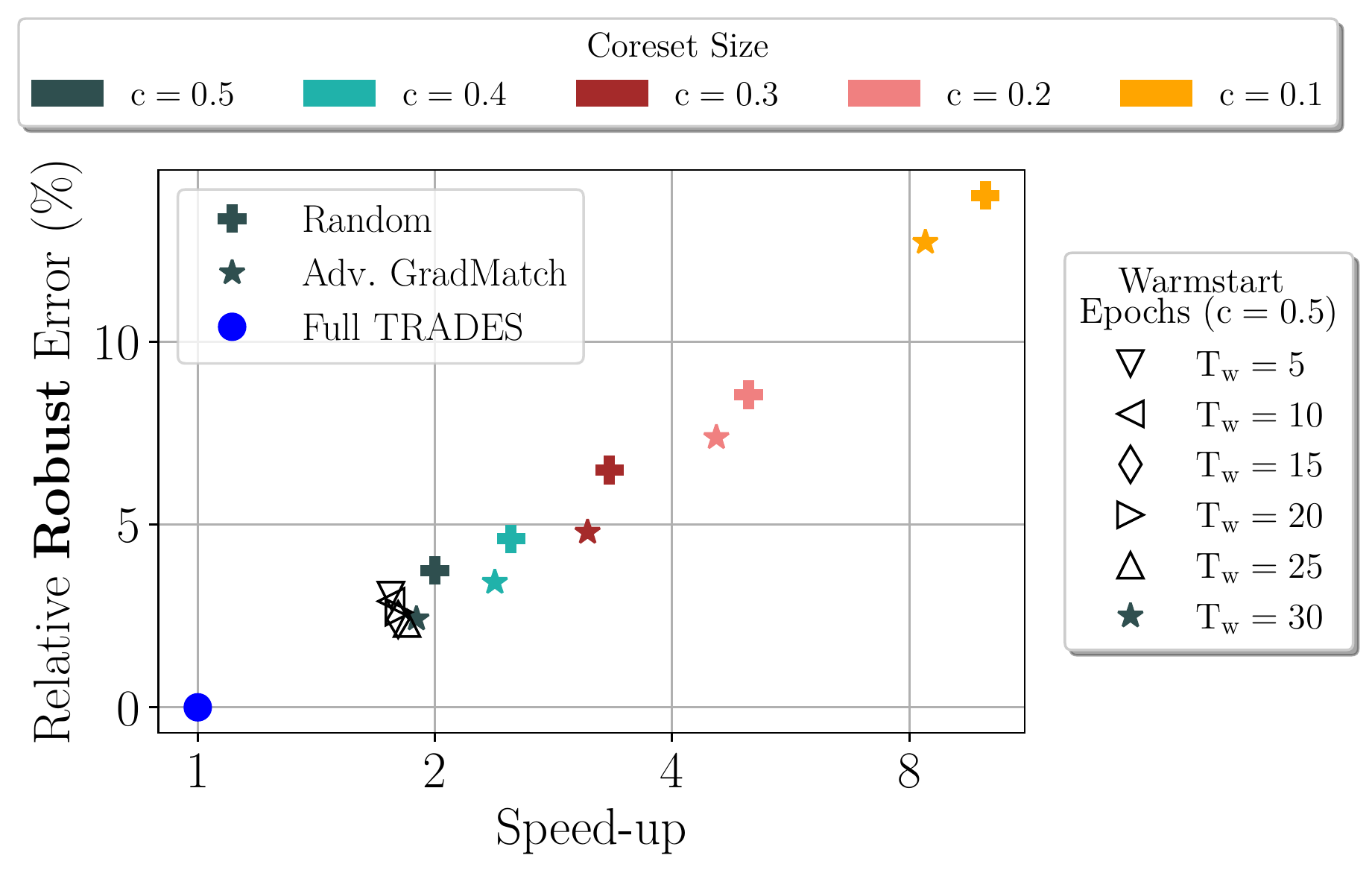}
                \caption{Relative robust error vs. speed up for TRADES. We compare our adversarial coreset selection (\textsc{GradMatch}) for a given subset size against random data selection.                     Furthermore, we show our results for a selection of different warm-start settings.}
			\label{fig:rand}
		\end{minipage}
\end{figure}

Moreover, we show that adversarial coreset selection gives a better approximation to $\ell_\infty$-PGD adversarial training compared to using FGSM~\cite{goodfellow2014explaining} as done in FAT~\cite{wong2020fast}.
To this end, we use our adversarial \textsc{GradMatch} to train neural networks with the original $\ell_\infty$-PGD objective.
We also train these networks using FAT~\cite{wong2020fast} that uses FGSM.
We train neural networks with a perturbation norm of $\norm{\varepsilon}_\infty \leq 8$.
Then, we evaluate the trained networks against PGD-50 adversarial attacks with different attack strengths to see how each network generalizes to unseen perturbations.
As seen in \Cref{fig:fat}, adversarial coreset selection is a closer approximation to $\ell_\infty$-PGD compared to FAT~\cite{wong2020fast}.
This indicates the success of the proposed approach in retaining the characteristics of the original objective as opposed to existing methods.

\subsection{Ablation Studies}
In this section, we perform a few ablation studies to examine the effectiveness of our adversarial coreset selection method.
First, we compare a random data selection with adversarial \textsc{GradMatch}.
\Cref{fig:rand} shows that for any given coreset size, our adversarial coreset selection method results in a lower robust error.
Furthermore, we modify the warm-start epochs for a fixed coreset size of 50\%.
As seen, the proposed method is not very sensitive to the number of warm-start epochs, although a longer warm-start is generally beneficial.
More experiments on the accuracy vs. speed-up trade-off and the importance of warm-start and batch-wise adversarial coreset selection can be found in \Cref{ap:extended}.

\section{Conclusion}\label{sec:conclusion}
In this paper, we proposed a general yet principled approach for efficient adversarial training based on the theory of coreset selection.
We discussed how repetitive computation of adversarial attacks for the entire training data could impede the training speed.
Unlike previous methods that try to solve this issue by making the adversarial attack more straightforward, here, we took an orthogonal path to reduce the training set size without modifying the attacker.
We drew a connection between greedy coreset selection algorithms and adversarial training using Danskin's theorem.
We then showed the flexibility of our adversarial coreset selection method by utilizing it for TRADES, $\ell_p$-PGD, and Perceptual Adversarial Training.
Our experimental results indicate that adversarial coreset selection can reduce the training time by more than 2-3 times with only a slight reduction in the clean and robust accuracy.

\noindent \textbf{Acknowledgements.} This research was undertaken using the LIEF HPC-GPG\-PU Facility hosted at the University of Melbourne. This Facility was established with the assistance of LIEF Grant LE170100200.
%
%

\begin{small}
\bibliographystyle{splncs04}
\bibliography{references}
\end{small}

\newpage
\appendix
\section{Greedy Subset Selection Algorithms}\label{ap:sec:greedy_selection}

This section briefly reviews the technical details of greedy subset selection algorithms used in our experiments.
Further details can be found in \cite{mirzasoleiman2020craig,killamsetty2021gradmatch}.

\subsection{\textsc{CRAIG}}\label{ap:sec:sec:craig}
As discussed previously, the goal of coreset selection is to find a subset of the training data such that the weighted gradient computed over this subset can give a good approximation to the full gradient.
Thus, \textsc{Craig}~\cite{mirzasoleiman2020craig} starts with explicitly writing down this objective as:
\begin{equation}\label{eq:gradient_app_craig}
    \argmin_{S \subseteq V}|S| ~\text{s.t.}~\max_{\boldsymbol{\theta}\in \boldsymbol{\Theta}} \norm{\sum_{i \in V} \nabla_{\boldsymbol{\theta}}\boldsymbol{\Phi} \left(\boldsymbol{x}_{i}, y_{i}; f_{\boldsymbol{\theta}}\right)-\sum_{j \in S} \gamma_{j}\nabla_{\boldsymbol{\theta}}\boldsymbol{\Phi} \left(\boldsymbol{x}_{j}, y_{j}; f_{\boldsymbol{\theta}}\right)} \leq \epsilon.
\end{equation}
Here, $V=\left[n\right]=\left\{1, 2, \dots, n\right\}$ denotes the training set.
The goal is to find a coreset $S \subseteq V$ and its associated weights $\gamma_{j}$ such that the objective of \Cref{eq:gradient_app_craig} is minimized.
To this end, Mirzasoleiman~\etal~\cite{mirzasoleiman2020craig} find an upper-bound on the gradient estimation error of \Cref{eq:gradient_app_craig}.
This way, it is shown that the coreset selection objective can be approximated by:
\begin{equation}\label{eq:craig_objective}
    S^{*}=\argmin_{S \subseteq V}|S|, \quad \text {s.t.} \quad L(S) \triangleq \sum_{i \in V} \min _{j \in S} d_{i j} \leq \epsilon,
\end{equation}
where
\begin{equation}\label{eq:craig_dij}
d_{i j} \triangleq \max _{\boldsymbol{\theta} \in \boldsymbol{\Theta}} \norm{ \nabla_{\boldsymbol{\theta}}\boldsymbol{\Phi} \left(\boldsymbol{x}_{i}, y_{i}; f_{\boldsymbol{\theta}}\right)-\nabla_{\boldsymbol{\theta}}\boldsymbol{\Phi} \left(\boldsymbol{x}_{j}, y_{j}; f_{\boldsymbol{\theta}}\right)}
\end{equation}
denotes the maximum pairwise gradient distances computed for all $i \in V$ and $j \in S$.
Then, Mirzasoleiman~\etal~\cite{mirzasoleiman2020craig} cast \Cref{eq:craig_objective} as the well-known \textit{submodular set cover problem} for which greedy solvers exist~\cite{minoux1978accelerated,nemhauser1978analysis,wolsey1982greedy}.

\subsection{\textsc{GradMatch}}\label{ap:sec:sec:gradmatch}
Killamsetty~\etal~\cite{killamsetty2021gradmatch} studies the convergence of adaptive data subset selection algorithms using \textit{stochastic gradient descent}~(SGD).
It is shown that the convergence bound consists of two terms: an irreducible noise-related term, and an additional gradient error term just like \Cref{eq:gradient_app_craig}.
Based on this analysis, Killamsetty~\etal~\cite{killamsetty2021gradmatch} propose to minimize this error directly.
To this end, they use the Orthogonal Matching Pursuit~(OMP)~\cite{pati1992omp,elenberg2016restricted} as their greedy solver, resulting in an algorithm called \textsc{Grad-Match}.
Since \textsc{GradMatch} minimizes the gradient estimation error given in \Cref{eq:gradient_app_craig} objective directly, it achieves a lower error compared to \textsc{Craig} that only minimizes an upper-bound of it.

\section{Further Details}
\subsection{Final Algorithm}\label{ap:final_alg}

\Cref{alg:adv_core} summarizes our adversarial coreset selection approach.
\begin{algorithm}[t!]
\vspace*{0.25em}
\caption{Adversarial Training with Coreset Selection~\label{alg:adv_core}} 
\begin{small}
\textbf{Input}: dataset $\mathcal{D}=\left\{\left(\boldsymbol{x}_{i}, y_{i}\right)\right\}_{i=1}^{n}$, neural network~${f_{\boldsymbol{\theta}}(\cdot)}$.\vspace*{0.25em}
\\
\textbf{Output}: robustly trained neural network~${f_{\boldsymbol{\theta}}(\cdot)}$.\vspace*{0.25em}
\\
\textbf{Parameters}: learning rate~$\alpha$, total epochs~$E$, warm-start coefficient~$\kappa$, coreset update period~$T$, batch size~$b$, coreset size~$k$, perturbation bound~$\varepsilon$.\vspace*{0.25em}
\begin{algorithmic}[1]
	\State Initialize~${\boldsymbol{\theta}}$ randomly.
	\State $\kappa_{\rm epochs} = \kappa \cdot E$
	\State $T_{\mathrm{warm}} = \kappa_{\rm epochs} \cdot k$
	\For {$t=1,2,\ldots, E$}
	\If {$t \leq T_{\mathrm{warm}}$}\vspace*{0.15em}
	\State $S \leftarrow \mathcal{D}$ \verb+\\ Warm-start with the entire data and uniform weights.+
	\ElsIf {$t \geq \kappa_{\rm epochs}$ \& $t\%T=0$} \vspace*{0.15em}
	\State $\mathcal{I} = \text{\textsc{BatchAssignments}}\left(\mathcal{D}, b\right)$ \verb+\\ Batch-wise selection.+\vspace*{0.15em}
	\State $\mathcal{Y} = \left\{{f_{\boldsymbol{\theta}}(\boldsymbol{x}_{i})}~|~\left(\boldsymbol{x}_{i}, y_{i}\right) \in \mathcal{D}\right\}$ \verb+\\ Computing the logits.+\vspace*{0.15em}
	\State $\mathcal{G} = \text{\textsc{AdvGradient}}\left(\mathcal{D}, \mathcal{Y}\right)$ \verb+\\ Using Eqs. 12 & 15 to find the gradients.+\vspace*{0.15em}
	\State $S\leftarrow\text{\textsc{GreedySolver}}\left(\mathcal{D}, \mathcal{I}, \mathcal{G}, \mathrm{coreset~size}=k\right)$ \verb+\\ Finding the coreset.+\vspace*{0.15em}
	\Else
	\State \textbf{Continue}
	\EndIf
	\For {$\mathrm{batch}$ in $S$}
	\State $\mathrm{batch_{adv}}=\text{\textsc{AdvExampleGen}}\left(\mathrm{batch}, f_{\boldsymbol{\theta}}, \varepsilon\right)$ \vspace*{0.15em}
	\State $\boldsymbol{\theta}\leftarrow\text{\textsc{SGD}}\left(\mathrm{batch_{adv}}, f_{\boldsymbol{\theta}}, \alpha\right)$ \verb+\\ Performing SGD over a batch of data.+ \vspace*{0.15em}
	\EndFor
	\EndFor
\end{algorithmic}
\end{small}
\end{algorithm}

\subsection{TRADES Gradient}\label{ap:trades_gradient}
To compute the \textit{second} gradient term in \Cref{eq:nn_gradeint_trades_penufinal} let us assume that $\boldsymbol{w}(\boldsymbol{\theta}) = f_{\boldsymbol{\theta}}(\boldsymbol{x}_{\rm{adv}})$ and $\boldsymbol{z}(\boldsymbol{\theta}) = f_{\boldsymbol{\theta}}(\boldsymbol{x})$.
We can write the aforementioned gradient as:
\begin{align}\label{eq:trades_chain_rule_ap}\nonumber
    \nabla_{\boldsymbol{\theta}} \mathcal{L}_{\mathrm{CE}}\left(f_{\boldsymbol{\theta}}(\boldsymbol{x}_{\rm{adv}}), f_{\boldsymbol{\theta}}(\boldsymbol{x})\right) 
    & = \nabla_{\boldsymbol{\theta}} \mathcal{L}_{\mathrm{CE}}\left(\boldsymbol{w}(\boldsymbol{\theta}), \boldsymbol{z}(\boldsymbol{\theta})\right)\\\nonumber
    &\stackrel{(1)}{=} \tfrac{\partial \mathcal{L}_{\mathrm{CE}}}{\partial \boldsymbol{w}} \nabla_{\boldsymbol{\theta}}\boldsymbol{w}(\boldsymbol{\theta}) + \tfrac{\partial \mathcal{L}_{\mathrm{CE}}}{\partial \boldsymbol{z}} \nabla_{\boldsymbol{\theta}}\boldsymbol{z}(\boldsymbol{\theta})\\\nonumber
    &\stackrel{(2)}{=} \nabla_{\boldsymbol{\theta}} \mathcal{L}_{\mathrm{CE}}\left(f_{\boldsymbol{\theta}}(\boldsymbol{x}_{\rm{adv}}), \text{\texttt{freeze}}\left(f_{\boldsymbol{\theta}}(\boldsymbol{x})\right)\right) \\ 
    &\quad + \nabla_{\boldsymbol{\theta}} \mathcal{L}_{\mathrm{CE}}\left(\text{\texttt{freeze}}\left(f_{\boldsymbol{\theta}}(\boldsymbol{x}_{\rm{adv}})\right), f_{\boldsymbol{\theta}}(\boldsymbol{x})\right).
\end{align}
Here, step (1) is derived using the multi-variable chain rule. 
Also, step (2) is the re-writing of step (1) by using the $\text{\texttt{freeze}}(\cdot)$ kernel that stops the gradients from backpropagating through its argument function.
Using this derivation, we can write the final TRADES gradient as:
\begin{align}\nonumber
	\nabla_{\boldsymbol{\theta}}\boldsymbol{\Phi} \left(\boldsymbol{x}, y; f_{\boldsymbol{\theta}}\right)
	= \nabla_{\boldsymbol{\theta}}\mathcal{L}_{\mathrm{CE}}\left(f_{\boldsymbol{\theta}}(\boldsymbol{x}), y\right)
	&+ \nabla_{\boldsymbol{\theta}} \mathcal{L}_{\mathrm{CE}}\left(f_{\boldsymbol{\theta}}(\boldsymbol{x}_{\rm{adv}}), \text{\texttt{freeze}}\left(f_{\boldsymbol{\theta}}(\boldsymbol{x})\right)\right)/\lambda \\ \label{eq:nn_gradeint_trades_final}
	&+ \nabla_{\boldsymbol{\theta}} \mathcal{L}_{\mathrm{CE}}\left(\text{\texttt{freeze}}\left(f_{\boldsymbol{\theta}}(\boldsymbol{x}_{\rm{adv}})\right), f_{\boldsymbol{\theta}}(\boldsymbol{x})\right)/\lambda.
\end{align}

\section{Implementation Details}\label{ap:sec:imp_det}

In this section, we provide the details of our experiments in \Cref{sec:experiments}.
We used a single NVIDIA Tesla V100-SXM2-16GB GPU for CIFAR-10~\cite{krizhevsky2009learning} and SVHN~\cite{netzer2011reading}, and a single NVIDIA Tesla V100-SXM2-32GB GPU for ImageNet-12~\cite{russakovsky2015imagenet,liu2020refool}.
Our implementation can be found on \href{https://github.com/hmdolatabadi/ACS}{GitHub}.

\subsection{Training Settings.}
\Cref{tab:hyper} shows the entire set of hyper-parameters and settings used for training the models of \Cref{sec:experiments}.

\subsection{Evaluation Settings}
For the evaluation of TRADES and $\ell_p$-PGD adversarial training, we use PGD attacks.
In particular, for TRADES and $\ell_\infty$-PGD adversarial training, we use $\ell_\infty$-PGD attacks with $\varepsilon=8/255$, step-size $\alpha=1/255$, 50 iterations, and 10 random restarts.
Also, for $\ell_2$-PGD adversarial training we use $\ell_2$-PGD attacks with $\varepsilon=80/255$, step-size $\alpha=8/255$, 50 iterations and 10 random restarts.

For Perceptual Adversarial Training~(PAT), we report the attacks' settings in \Cref{tab:hyper_PAT}.
We evaluated each case using the same set of unseen/seen attacks as in Laidlaw~\etal~\cite{laidlaw2021pat}.
However, since we trained our model with slightly different $\varepsilon$ bounds, we changed the attacks' settings accordingly.

\begin{table}[htp]
	\caption{Hyper-parameters of the attacks used for the evaluation of PAT models.}
	\label{tab:hyper_PAT}
	\begin{center}
		\begin{small}
			\setlength\tabcolsep{1.6pt}
			\begin{tabular}{cccc}
				\toprule
				\textbf{\scriptsize{Dataset}}
				& \textbf{Attack}
				& \textbf{Bound}
				& \textbf{Iterations} \\
				\midrule
				\parbox[t]{2mm}{\multirow{6}{*}{\rotatebox[origin=c]{90}{CIFAR-10}}}
				& AutoAttack-$\ell_2$~\cite{croce2020autoattack}                   & $1$	           & $20$ \\
				& AutoAttack-$\ell_\infty$~\cite{croce2020autoattack}              & $8/255$	       & $20$ \\
				& StAdv~\cite{xiao2018stadv}                                       & $0.02$	       & $50$ \\
				& ReColor~\cite{laidlaw2019recolor}                                & $0.06$	       & $100$ \\
				& PPGD~\cite{laidlaw2021pat}                                       & $0.40$	       & $40$ \\
				& LPA~\cite{laidlaw2021pat}                                        & $0.40$	       & $40$ \\                     
				\midrule
				\parbox[t]{2mm}{\multirow{7}{*}{\rotatebox[origin=c]{90}{ImageNet-12}}}
				& AutoAttack-$\ell_2$~\cite{croce2020autoattack}                   & $1200/255$	   & $20$ \\
				& AutoAttack-$\ell_\infty$~\cite{croce2020autoattack}              & $4/255$	       & $20$ \\
				& JPEG~\cite{kang2019JPEG}                                         & $0.125$	       & $200$ \\
				& StAdv~\cite{xiao2018stadv}                                       & $0.02$	       & $50$ \\
				& ReColor~\cite{laidlaw2019recolor}                                & $0.06$	       & $200$ \\
				& PPGD~\cite{laidlaw2021pat}                                       & $0.35$	       & $40$ \\
				& LPA~\cite{laidlaw2021pat}                                        & $0.35$	       & $40$ \\
				\bottomrule
			\end{tabular}
		\end{small}
	\end{center}
\end{table}

\begin{sidewaystable}[htp]
	\caption{Training details for experimental results of \Cref{sec:experiments}.}
	\label{tab:hyper}
	\begin{center}
		\begin{scriptsize}
			\begin{tabular}{lcccccc}
				\toprule
				\multirow{2}{*}{\textbf{Hyper-parameter}}    & \multicolumn{6}{c}{\textbf{Experiment}}\\
				\cmidrule{2-7}
				& TRADES                                    & $\ell_\infty$-PGD                & $\ell_2$-PGD                    & Fast-LPA                   & Fast-LPA                  & Fast Adv.\\
				\midrule
				\textbf{Dataset}                            & CIFAR-10                         & CIFAR-10                        & SVHN                       & CIFAR-10                  & ImageNet-12 & CIFAR-10\\
				\textbf{Model Arch.}                        & ResNet-18                        & ResNet-18                       & ResNet-18                  & ResNet-50                 & ResNet50    & ResNet-18\\
				\midrule
				\textbf{Optimizer}                          & SGD                              & SGD                             & SGD                        & SGD                       & SGD         & SGD\\
				\textbf{Scheduler}                          & Multi-step                       & Multi-step                      & Multi-step                 & Multi-step                & Multi-step  & Multi-step\\
				\textbf{Initial lr.}                        & $0.1$                            & $0.01$                          & $0.1$                      & $0.1$                     & $0.1$       & $0.1$\\
				\textbf{lr. Decay (epochs)}                 & $0.1$ ($75$, $90$)               & $0.1$ ($80$, $100$)       & $0.1$ ($75$, $90$, $100$)  & $0.1$ ($75$, $90$, $100$) & $0.1$ ($45$, $60$, $80$)) & $0.1$ ($37$, $56$)\\
				\textbf{Weight Decay}                       & $2\cdot10^{-4}$                  & $5\cdot10^{-4}$                 & $5\cdot10^{-4}$            & $2\cdot10^{-4}$           & $2\cdot10^{-4}$ & $5\cdot10^{-4}$\\
				\textbf{Batch Size (full)}                  & $128$                            & $128$                           & $128$                      & $50$                      & $50$      & $128$\\
				\textbf{Total Epochs}                       & $100$                            & $120$                           & $120$                      & $120$                     & $90$      & $60$\\
				\midrule
				\textbf{Coreset Size}                       & $50$\%                           & $50$\%                          & $30$\%                     & $40$\%                     & $50$\%   & $50$\%\\
				\textbf{Coreset Batch Size}                 & $20$                             & $20$                            & $20$                       & $20$                       & $20$     & $20$\\
				\textbf{Warm-start Epochs}                  & $30$                             & $36$                            & $22$                       & $29$                       & $27$     & $22$\\
				\textbf{Coreset Selection Period (epochs)}  & $20$                             & $20$                            & $20$                       & $10$                       & $15$     & $5$\\
				\midrule
				\textbf{Visual Similarity Measure}            & $\ell_\infty$                  & $\ell_\infty$                   & $\ell_2$                   & LPIPS (AlexNet~\cite{krizhevsky2012alexnet}) & LPIPS (AlexNet) & $\ell_\infty$\\
				\textbf{$\varepsilon$ (Bound on Visual Sim.)} & $8/255$                        & $8/255$                         & $80/255$                   & $0.5$                      & $0.25$   & $8/255$\\
				\textbf{Attack Iterations (Training)}         & $10$                           & $10$                            & $10$                       & $10$                       & $10$     & $1$\\
				\textbf{Attack Iterations (Coreset Selection)}& $10$                           & $1$                             & $10$                       & $10$                       & $10$     & $1$\\
				\textbf{Attack Step-size}                     & $1.785/255$                     & $1.25/255$                      & $8/255$                    & -                          & -        & $10/255$\\      
				\bottomrule
			\end{tabular}
		\end{scriptsize}
	\end{center}
\end{sidewaystable}

\section{Extended Experimental Results}\label{ap:extended}

\subsection{Extended Results of PAT vs. Unseen Attacks}
\Cref{tab:LPA} shows the full details of our experiments on PAT~\cite{laidlaw2021pat}.
In each case, we train ResNet-50~\cite{he2016deep} classifiers using LPIPS~\cite{zhang2018lpips} objective of PAT~\cite{laidlaw2021pat}.
All the training hyper-parameters are fixed.
The only difference is that we enable adversarial coreset selection as our method.
During inference, we evaluate each trained model against a few unseen attacks, as well as two variants of Perceptual Adversarial Attacks~\cite{laidlaw2021pat} that the models are trained initially on.
As can be seen, adversarial coreset selection can significantly reduce the training time while experiencing only a tiny reduction in the average robust accuracy.

\begin{sidewaystable}[p!]
	\caption{Clean and robust accuracy (\%) and total training time (mins) of Perceptual Adversarial Training for CIFAR-10 and ImageNet-12 datasets. The training objective uses Fast Lagrangian Perceptual Attack (LPA)~\cite{laidlaw2021pat} to train the network. At test time, the networks are evaluated against attacks not seen during training and different versions of Perceptual Adversarial Attack (PPGD and LPA). In each dataset, the unseen attacks were selected similar to Laidlaw~\etal~\cite{laidlaw2021pat}. Please see the \Cref{ap:sec:imp_det} for more information about the settings.}
	\label{tab:LPA}
	\begin{center}
		\begin{scriptsize}
			\setlength\tabcolsep{1.5pt}
			\def\arraystretch{2.5}
			\begin{tabular}{cccccccccccc}
				\toprule
				\parbox[t]{2mm}{\multirow{2}{*}{\rotatebox[origin=c]{90}{\textbf{\scriptsize{Dataset}}}}}
				&\multirow{2}{*}{\textbf{Training Method}}
				& \multirow{2}{*}{\textbf{Clean}}
				& \multicolumn{6}{c}{\textbf{Unseen Attacks}}
				& \multicolumn{2}{c}{\textbf{Seen Attacks}} 
				& \multirow{2}{*}{\shortstack{\textbf{Train. Time}\\(mins)}}\\
				\cmidrule(lr){4-9} \cmidrule(lr){10-11}
				& & & Auto-$\ell_2$~\cite{croce2020autoattack}   & Auto-$\ell_\infty$~\cite{croce2020autoattack}      & JPEG~\cite{kang2019JPEG}    & StAdv~\cite{xiao2018stadv} 
				& ReColor\cite{laidlaw2019recolor}   & Mean    & PPGD     & LPA      & \\
				\midrule
				\parbox[t]{2mm}{\multirow{3}{*}{\rotatebox[origin=c]{90}{\scriptsize{\textbf{CIFAR-10}}}}}
				& Adv. \textsc{Craig} (Ours)             & $\mathbf{83.21}$        & $39.98$	& $33.94$	       &  -           & $49.60$	 & $62.69$	  & $\mathbf{46.55}$  & $19.56$	& $7.42$       & $\mathbf{767.34}$ \\
				& Adv. \textsc{GradMatch} (Ours)         & $\mathbf{83.14}$	       & $39.20$	& $34.11$	       &  -           & $48.86$	 & $62.26$	  & $\mathbf{46.11}$   & $19.94$	& $7.54$   & $\mathbf{787.26}$ \\
				& Full PAT (Fast-LPA)                    & $\mathbf{86.02}$	       & $43.27$	& $37.96$	       &  -           & $48.68$	 & $62.23$	  & $\mathbf{48.04}$   & $22.62$	& $8.01$   & $\mathbf{1682.94}$ \\
				\midrule
				\parbox[t]{2mm}{\multirow{3}{*}{\rotatebox[origin=c]{90}{\tiny{\textbf{ImageNet-12}}}}}
				& Adv. \textsc{Craig} (Ours)             & $\mathbf{86.99}$        & $51.54$	& $60.42$	       & $71.79$      & $37.47$	 & $44.04$	  & $\mathbf{53.05}$    & $29.04$    & $14.07$   & $\mathbf{2817.06}$ \\
				& Adv. \textsc{GradMatch} (Ours)         & $\mathbf{87.08}$	       & $51.38$	& $60.64$          & $72.15$      & $35.83$	 & $45.83$	  & $\mathbf{53.17}$    & $28.36$    & $13.11$   & $\mathbf{2865.72}$ \\
				& Full PAT (Fast-LPA)                    & $\mathbf{91.22}$	       & $57.37$    & $66.89$          & $76.25$      & $19.29$	 & $46.35$	  & $\mathbf{53.23}$    & $33.17$    & $13.49$   & $\mathbf{5613.12}$ \\
				\bottomrule
			\end{tabular}
		\end{scriptsize}
	\end{center}
\end{sidewaystable}

\subsection{Trade-offs}
\begin{figure*}[p!]
	\centering
	\begin{subfigure}{.49\textwidth}
		\centering
		\includegraphics[width=1.0\textwidth]{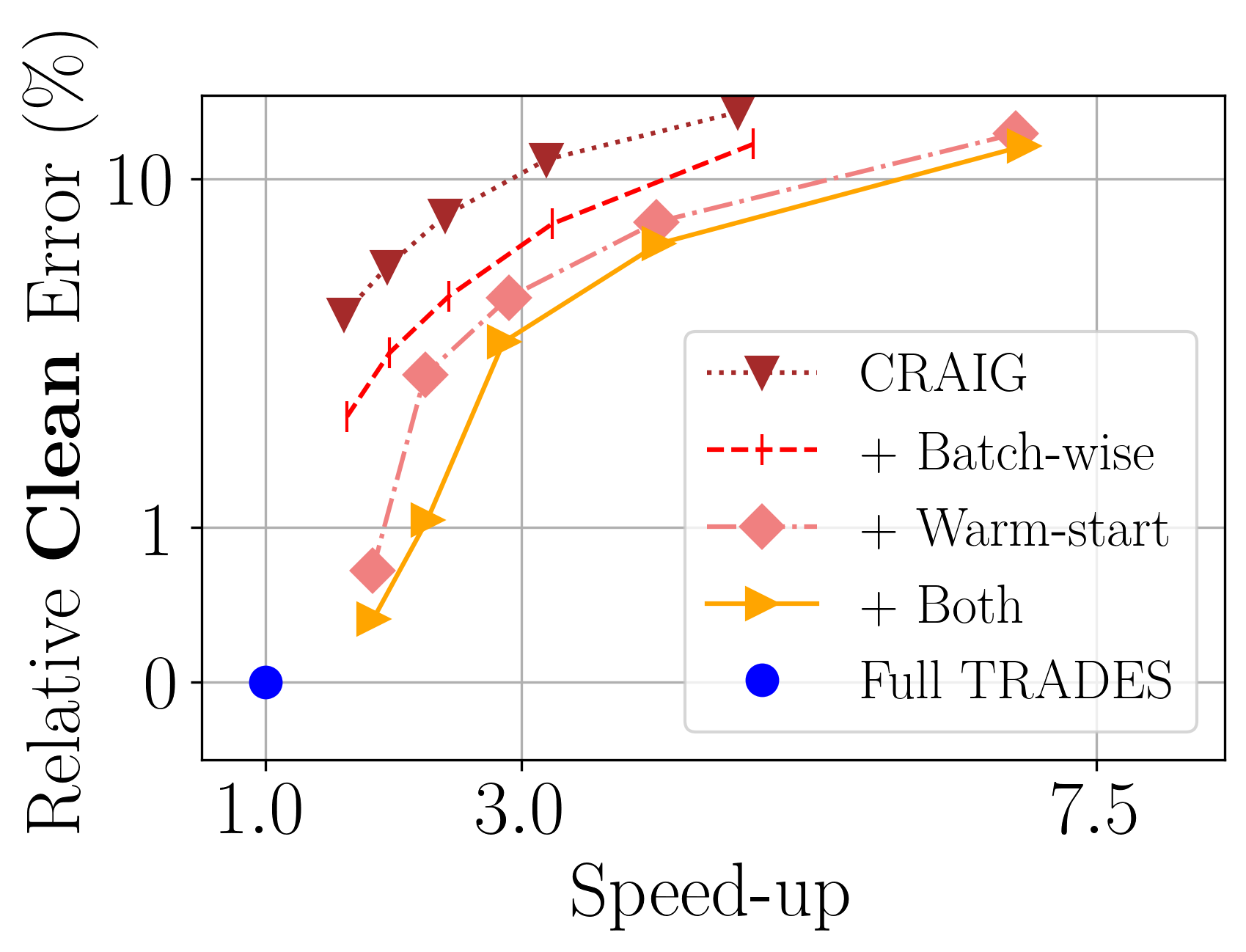}
		\caption{}
		\label{fig:tradeoff:craig_clean}
	\end{subfigure}
	\begin{subfigure}{.49\textwidth}
		\centering
		\includegraphics[width=1.0\textwidth]{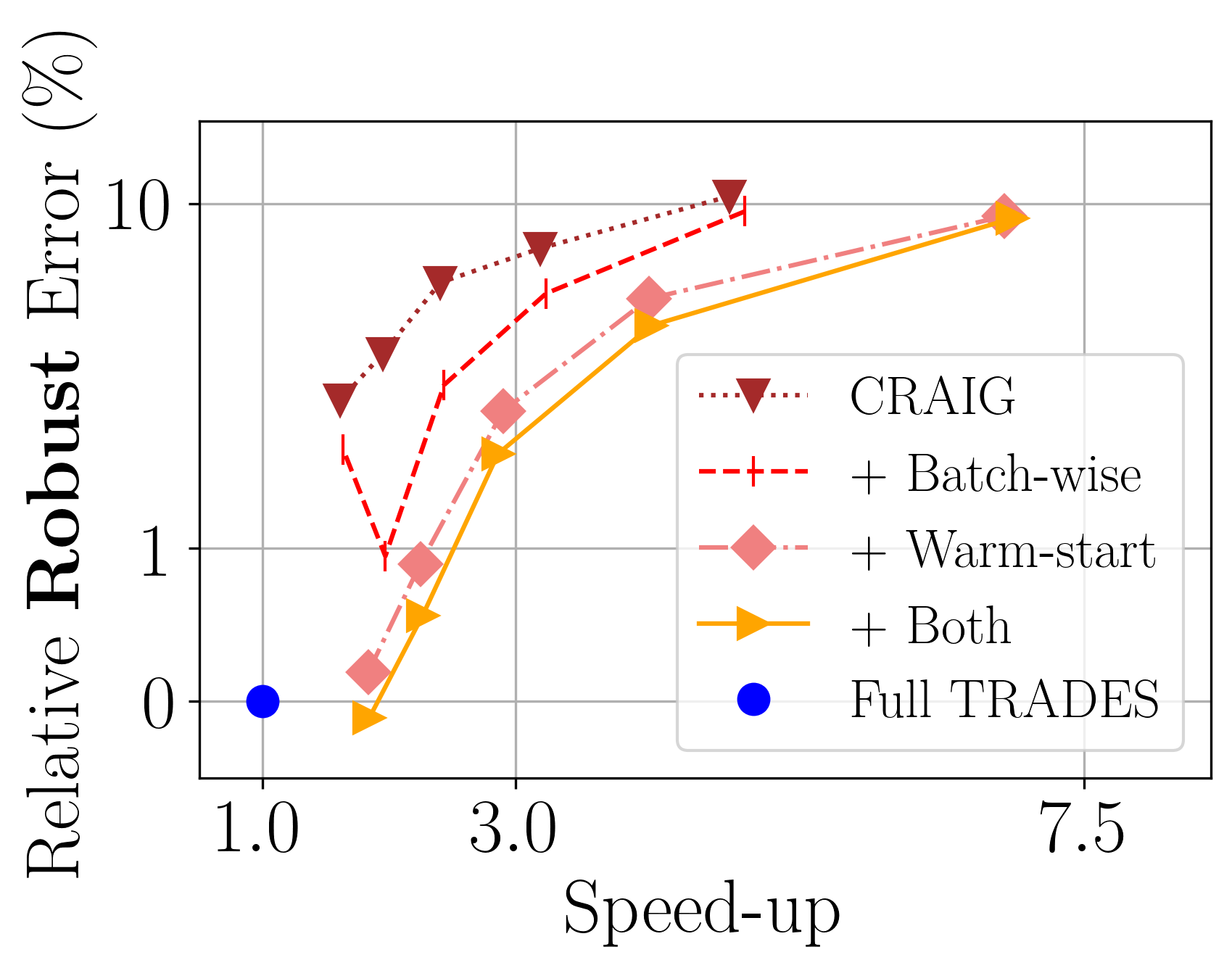}
		\caption{}
		\label{fig:tradeoff:craig_robust}
	\end{subfigure}\\
	\begin{subfigure}{.49\textwidth}
		\centering
		\includegraphics[width=1.0\textwidth]{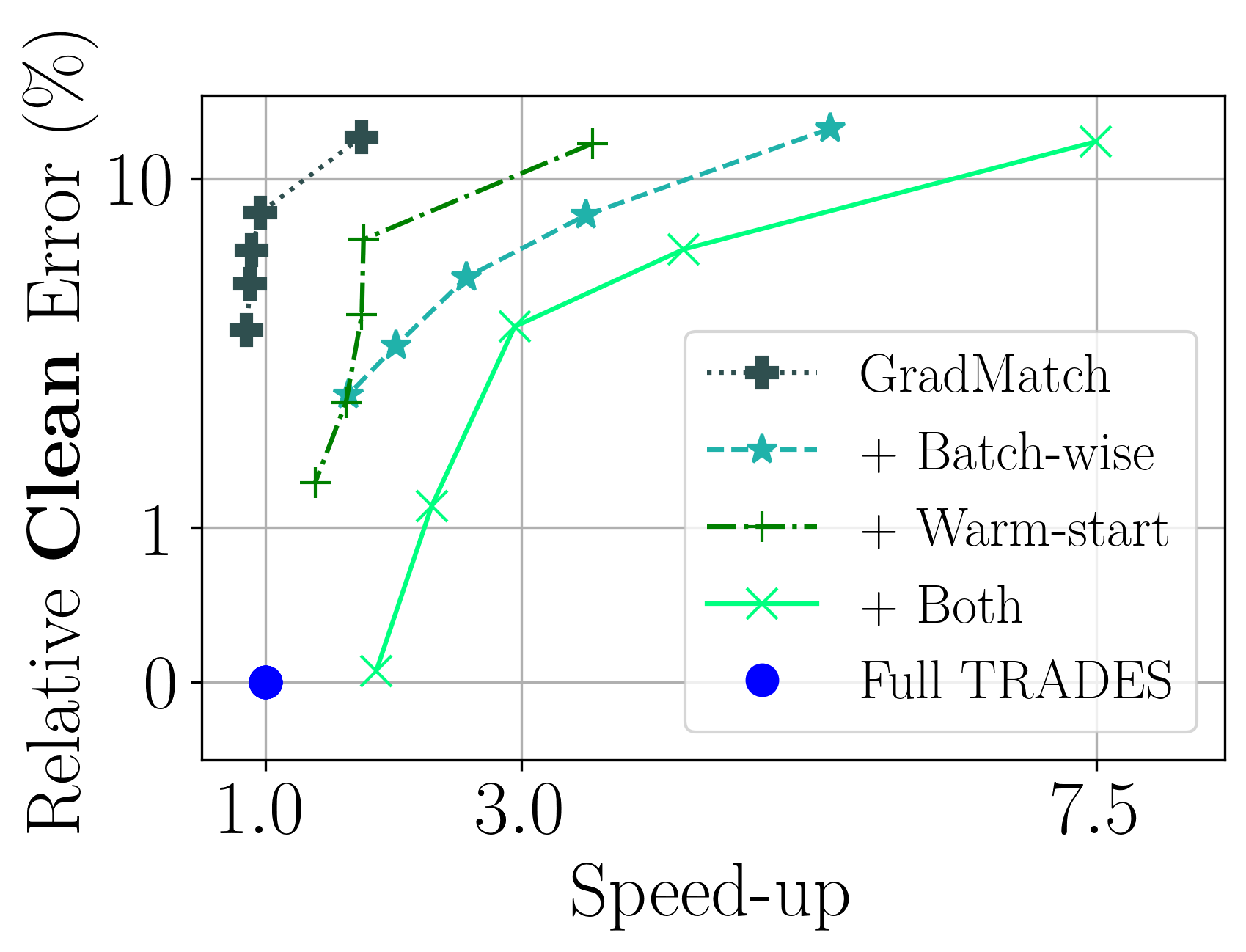}
		\caption{}
		\label{fig:tradeoff:gradmatch_clean}
	\end{subfigure}
	\begin{subfigure}{.49\textwidth}
		\centering
		\includegraphics[width=1.0\textwidth]{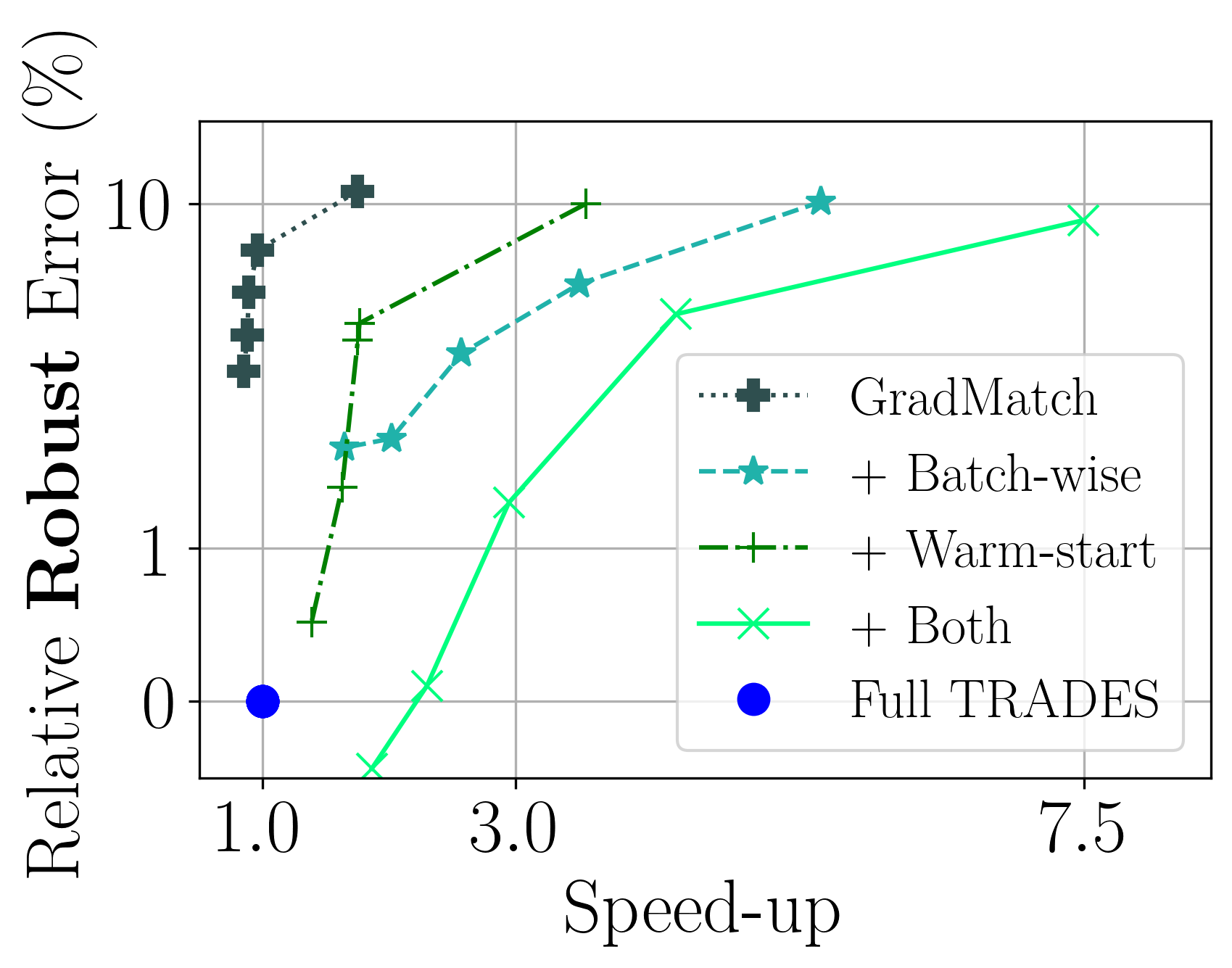}
		\caption{}
		\label{fig:tradeoff:gradmatch_robust}
	\end{subfigure}
	\caption{Relative error vs. speed up curves for different versions of adversarial coreset selection in training CIFAR-10 models using the TRADES objective. In each figure, the coreset size is changed from 50\% to 10\% (left to right). (a, b) Clean and robust error vs. speed up compared to full TRADES for different versions of adversarial \textsc{Craig}. (c, d) Clean and robust error vs. speed up compared to full TRADES for different versions of adversarial \textsc{GradMatch}.}
	\label{fig:tradeoff}
\end{figure*}

Here, we study the accuracy vs. speed-up trade-off in adversarial coreset selection.
For this study, we train our adversarial coreset selection method using different versions of \textsc{Craig}~\cite{mirzasoleiman2020craig} and \textsc{GradMatch}~\cite{killamsetty2021gradmatch} on CIFAR-10 using TRADES.
In particular, for each method, we start with the base algorithm and add the batch-wise selection and warm-start one by one.
Also, to capture the effect of the coreset size, we vary this number from 50\% to 10\% in each case.
\Cref{fig:tradeoff} shows the clean and robust error vs. speed-up compared to full adversarial training.
In each case, the combination of warm-start and batch-wise versions of the adversarial coreset selection gives the best performance.
Moreover, the training speed increases as we gradually decrease the coreset size.
However, this gain in training speed is achieved at the cost of increasing the clean and robust error.
Both of these observations are in line with that of Killamsetty~\etal~\cite{killamsetty2021gradmatch} around vanilla coreset selection.

\subsection{Training with a Mixture of Coreset and Non-coreset Data}
In this section, we run an experiment similar to that of Tsipras~\etal~\cite{tsipras2019robustness}.
Specifically, we minimize the average of adversarial and vanilla training in each epoch.
The non-coreset data is treated as clean samples to minimize the vanilla objective, while for the coreset samples, we would perform adversarial training.
\Cref{tab:HH_PGD} shows the results of this experiment.
As seen, adding the non-coreset data as clean inputs to the training improves the clean accuracy while decreasing the robust accuracy.
These results align with the observations of Tsipras~\etal~\cite{tsipras2019robustness} around the existence of a trade-off between clean and robust accuracy.

\begin{sidewaystable}[p!]
		\caption{Performance of $\ell_\infty$-PGD over CIFAR-10. In ``Half-Half'', we mix half adversarial coreset selection samples with another half of clean samples and train a neural network similar to~\cite{tsipras2019robustness}.
		In ``ONLY-Core'' we just use adversarial coreset samples. Settings are given in \Cref{tab:hyper}.
		The results are averaged over 5 runs.}
		\label{tab:HH_PGD}
		\begin{center}
			\begin{scriptsize}
				\setlength\tabcolsep{0.45em}
				\def\arraystretch{1.1}
				\begin{tabular}{cccccccc}
					\toprule
					\multirow{2}{*}{\textbf{Training Method}}  &  \multicolumn{2}{c}{\textuparrow~\textbf{Clean} (\%)}     & \multicolumn{2}{c}{\textuparrow~\textbf{RACC} (\%)} & \multicolumn{2}{c}{\textdownarrow~\textbf{T} (mins)}\\
					\cmidrule(lr){2-3}\cmidrule(lr){4-5}\cmidrule(lr){6-7}
					                                    & \textbf{ONLY Core} & \textbf{Half-Half}                   & \textbf{ONLY Core} & \textbf{Half-Half}             & \textbf{ONLY Core} & \textbf{Half-Half}\\
					\midrule
					Adv. \textsc{Craig}                 & $80.36$    & $84.43$                                      & $45.07$  & $39.83$         & $148.01$ & $152.34$\\
					Adv. \textsc{GradMatch}             & $80.67$    & $84.31$                                      & $45.23$  & $40.05$         & $148.03$ & $153.18$\\
					Full Adv. Training                  & \multicolumn{2}{c}{$83.14$}                               & \multicolumn{2}{c}{$41.39$}  & \multicolumn{2}{c}{$292.87$}     \\
					\bottomrule
				\end{tabular}
			\end{scriptsize}
		\end{center}
\end{sidewaystable}

\end{document}